\documentclass[10pt,twocolumn,letterpaper]{article}

\usepackage[pagenumbers]{cvpr} 

\definecolor{cvprblue}{rgb}{0.21,0.49,0.74}
\usepackage[pagebackref,breaklinks,colorlinks,allcolors=cvprblue]{hyperref}
\usepackage[accsupp]{axessibility}  

\usepackage{tabularx}
\usepackage{longtable}
\usepackage{multirow}
\usepackage{booktabs} 
\usepackage{caption}
\usepackage{graphicx}
\usepackage{subcaption}
\usepackage{float}
\usepackage{placeins}
\usepackage{tabulary,multirow}
\usepackage{ragged2e} 
\usepackage{url}
\newcolumntype{Y}{>{\centering\arraybackslash}X}

\title{Scaling Pre-training to One Hundred Billion Data for Vision Language Models}

\author{
    Xiao Wang \hspace{0.8em} Ibrahim Alabdulmohsin \hspace{0.8em} Daniel Salz \hspace{0.8em} Zhe Li \hspace{0.8em} Keran Rong\thanks{K. Rong is now at xAI; X. Zhai is now at Meta.} \hspace{0.8em} Xiaohua Zhai\footnotemark[1] \\
    Google DeepMind \quad {\tt\small \{wangxiao, ibomohsin\}@google.com}
}

\begin{document}
\maketitle

\begin{abstract}
We provide an empirical investigation of the potential of pre-training vision-language models on an unprecedented scale: 100 billion examples. We find that model performance tends to saturate at this scale on many common Western-centric classification and retrieval benchmarks, such as COCO Captions. Nevertheless, tasks of cultural diversity achieve more substantial gains from the 100-billion scale web data, thanks to its coverage of long-tail concepts. Furthermore, we analyze the model's multilinguality and show gains in low-resource languages as well. In addition, we observe that reducing the size of the pretraining dataset via quality filters like using CLIP, typically used to enhance performance, may inadvertently reduce the cultural diversity represented in large-scale datasets. Our results highlight that while traditional benchmarks may not benefit significantly from scaling noisy, raw web data to 100 billion examples, this data scale is vital for building truly inclusive multimodal systems.
\end{abstract}

\begin{figure*}[t!]
\begin{subfigure}{0.23\textwidth}
    \includegraphics[width=\columnwidth]{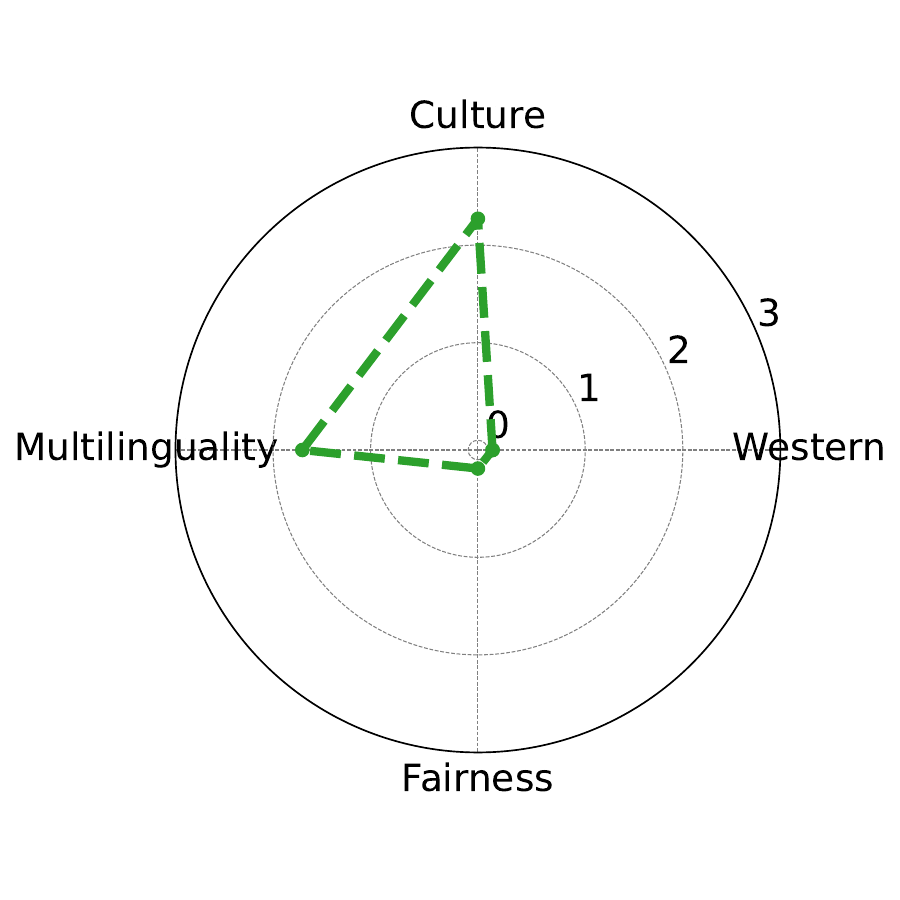}
\end{subfigure}
\hfill
\begin{subfigure}{0.74\textwidth}
\vspace{-0em}\includegraphics[width=0.24\columnwidth]{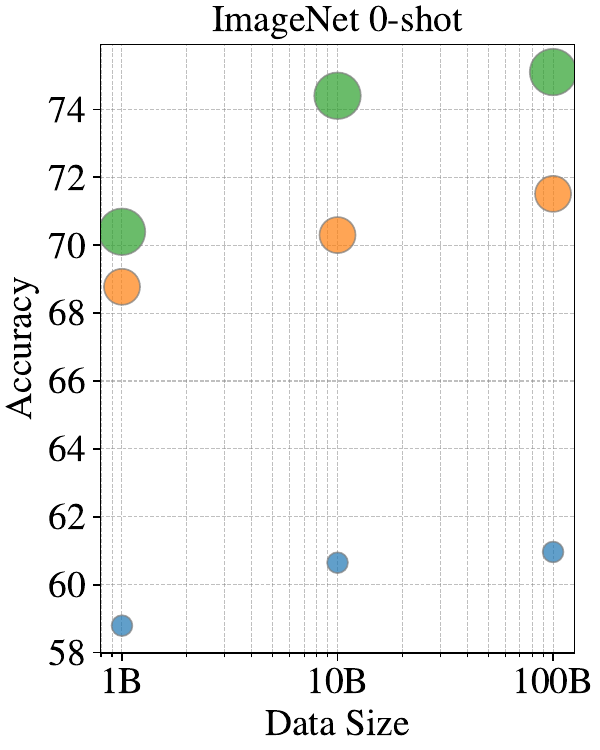}
\includegraphics[width=0.24\columnwidth]{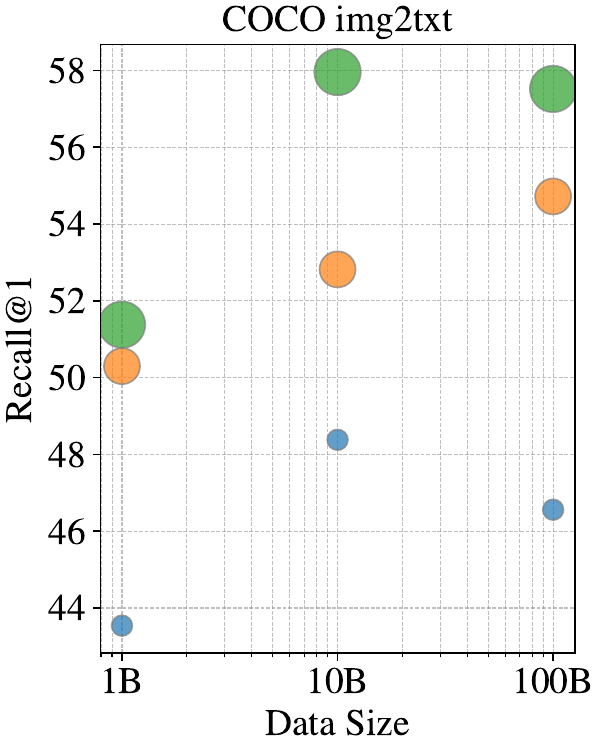}
\includegraphics[width=0.24\columnwidth]{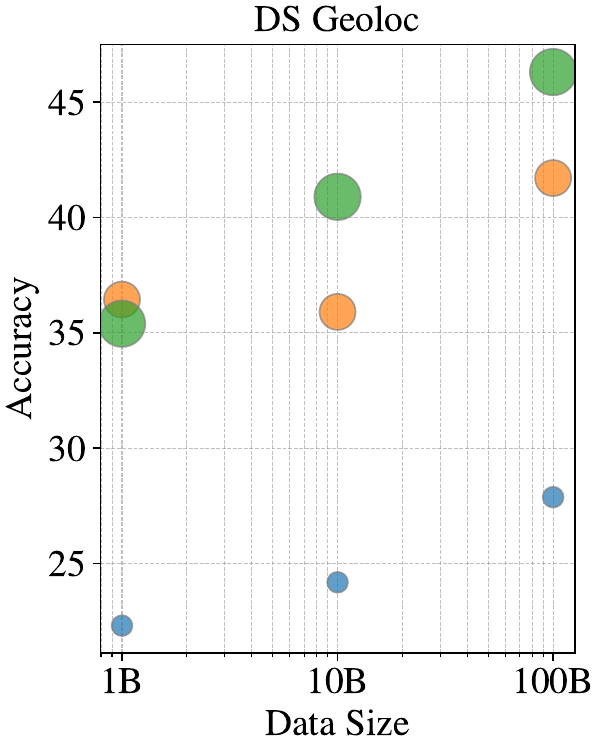}
\includegraphics[width=0.24\columnwidth]{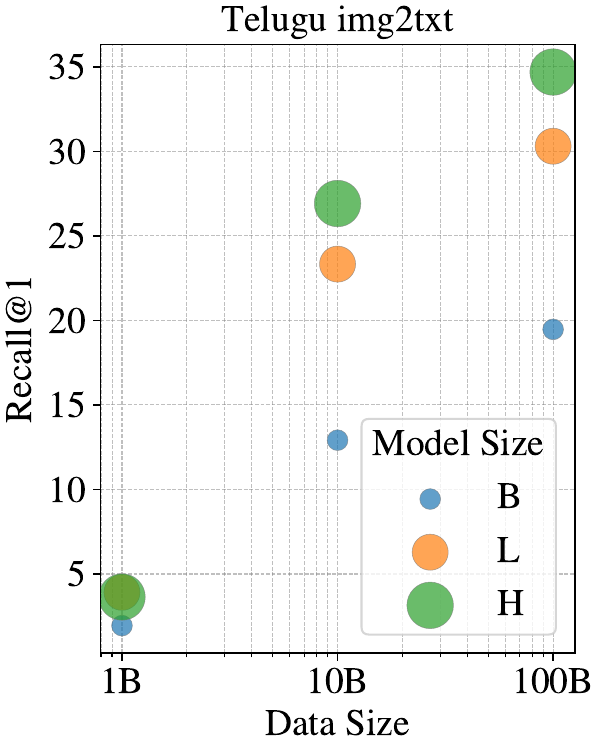}
\end{subfigure}
    \caption{{\sc left:} Scaling the data from 10B to 100B examples enhances cultural diversity and multilingual capabilities more prominently than other metrics. The numbers in the radar plot represent the improvement in accuracy in absolute terms for ViT-H/14 when data scale is increased from 10B to 100B examples, averaged across all tasks. {\sc right:} Illustrative examples of the impact of data scale. The leftmost two are Western-centric metrics, which benefit little by scaling the data to 100B, while the rightmost two are illustrative of cultural diversity and multilinguality. The language Telugu, for example, makes up  $<0.04\%$ of the web and benefits a lot from the 100B data scale.}
    \label{fig:main}
\end{figure*}

\section{Introduction}
Progress in vision-language models (VLMs) has been tied to the availability of large-scale datasets. Large datasets fuel the development of powerful models, which are capable of understanding and generating complex relationships between images and text. In turn, such models have pushed boundaries in tasks like zero-shot image classification, image captioning and visual question answering. 

This relationship between data scale and model performance often follows a power law $f(x)=\alpha\, x^{-c}+\varepsilon$, where $f(x)$ is a model performance metric such as its error rate and $x$ is the data size~\citep{mukherjee2003estimating,hestness2017deep,johnson-etal-2018-predicting,rosenfeld2019constructive,kaplan2020scaling,ghorbani2021scaling,alabdulmohsin2022revisiting,bansal2022data,zhai2106scaling}. These ``scaling laws,'' as they came to be known in the literature, have been used, among others, to determine the training data size needed to achieve a specified level of accuracy~\citep{cho2015much,beleites2013sample,figueroa2012predicting} and to optimize the model size~\citep{kaplan2020scaling,hoffmann2022training,alabdulmohsin2024getting}. They have also been justified theoretically using space-partitioning arguments~\citep{bahri2021explaining,hutter2021learning,sharma2022scaling}. Importantly, a power law implies that increasing the amount of training data can yield diminishing, but \emph{still worthwhile}, returns in terms of accuracy and capability. 

Driven by these potential benefits, the field has witnessed a concerted effort towards scaling up the size of vision-language datasets. Early works focused on web curated datasets like Conceptual Captions~\citep{sharma2018conceptual}, which provided millions of image-caption pairs for pre-training~\citep{sharma2018conceptual}. Subsequent work leveraged large-scale web crawling to create even larger datasets. In particular, the Common Crawl project~\citep{commoncrawl2021}---a repository of publicly available web data---became a foundational resource for constructing many of these web-scale datasets. From this foundation emerged datasets like LAION-400M/2B/5B~\citep{schuhmann2022laion}, DataComp~\citep{datacomp}, WebLI~\citep{chen2022pali} and Multimodal C4~\citep{zhu2024multimodal}, pushing the boundaries of dataset size to billions  of image-text pairs, thereby accelerating progress in VLMs. This is similar to how ImageNet~\citep{deng2009imagenet},  JFT-300M~\citep{sun2017revisiting}--a dataset of 300 million images with noisy labels--and its larger variant JFT-3B~\citep{zhai2106scaling} accelerated progress in supervised image pre-training previously.

Despite these advancements, the largest reported datasets to date have plateaued at around 10 billion image-text pairs. This raises the question: \emph{what further benefits are unlocked by pushing the data scale by one order of magnitude to 100 billion unique examples?}

To answer this question, we develop WebLI-100B, a novel dataset containing 100 billion image-text pairs, representing a \emph{tenfold} increase over the largest reported vision-language datasets. To recall, the original WebLI dataset contains 10 billion examples and has been instrumental in training state-of-the-art models like PaliGemma~\citep{beyer2024paligemma,steiner2024paligemma2familyversatile} and SigLIP~\citep{zhai2023sigmoidlosslanguageimage}, and influenced the development of other research directions, such as mitigating social biases~\citep{alabdulmohsin2024clip}, improving cultural diversity~\citep{pouget2024no}, and scaling open-vocabulary object detection~\citep{minderer2024scaling}.
In this work, our primary goal is to provide an empirical investigation to the impact of this data scale on a range of downstream tasks and, importantly, to explore aspects beyond traditional performance metrics. 
For example, when applied to geo-localization tasks based on Dollar Street~\citep{rojas2022dollar}---a metric for evaluating cultural diversity---ViT-L/16 trained on a single epoch of 100 billion data achieves an accuracy of 41.7\%. By contrast, the same model trained on ten epochs of 10 billion data achieves an accuracy of 35.9\% only, despite both models using the same amount of training compute. We attribute these gains, in part, to the dataset's ability to capture a wider range of long-tail cultural concepts that require a substantial data size to become salient. Furthermore, data scaling also enhances multilinguality, leading to an improvement in low-resource languages. Figure~\ref{fig:main} summarizes the  improvements achieved through data scaling.

\paragraph{Statement of Contribution.} Our major contributions can be summarized as follows:
\begin{itemize}
    \item We conduct a rigorous, large-scale investigation that answers a core, unanswered question in VLMs: What is the true impact of scaling pre-training data to an unprecedented 100 billion unique examples?
    \item We provide the first empirical evidence that scaling to 100 billion unfiltered examples yields diminishing returns on traditional Western-centric benchmarks, while significantly enhancing inclusivity across culturally diverse and long-tail domains. This discovery offers valuable guidance for optimizing resource allocation and underscores that inclusive evaluation is a necessity to capture the full value of data scaling.
    \item We identify a critical trade-off in CLIP-based data curation: while quality filters effectively clean data, they disproportionately exclude cultural contexts, thereby diminishing dataset diversity. Notably, we show that this bias does not vanish with data scaling, which remains significant even at the magnitude of 100 billion examples.
    \item We show that language imbalance persists at this magnitude. By upsampling under-represented languages, we significantly improve low-resource and cultural diversity benchmarks with minimal impact on high-resource performance.
    \item In terms of fairness, we show that scaling to 100 billion examples improves error disparity, closely related to Equaized Odds, but has no effect on representation and association biases.
\end{itemize}

\section{Related Work}\label{sect:related_work}

\paragraph{Data Scaling.}

The study of scaling laws in LLMs has become a critical area of research in NLP. \cite{hestness2017deep} and \cite{kaplan2020scaling} were among the first to explore the relationship between model size, dataset size, and compute, demonstrating predictable power-law scaling of performance. \cite{henighan2020scaling} further emphasized the crucial role of data, showing that substantial performance gains can be achieved by increasing the size and quality of the training dataset, even with fixed model size. DeepMind's Chinchilla~\citep{hoffmann2022training} provided compelling evidence for this data-centric approach, demonstrating that smaller models trained on much larger datasets can achieve comparable performance to larger models, given the same computational budget, therefore shifting the focus towards optimizing the scale of  data.

In computer vision, early works, such as ImageNet~\citep{deng2009imagenet}, demonstrated the profound impact of dataset size and diversity on model generalization. Subsequent efforts like JFT-300M~\citep{sun2017revisiting} emphasized the importance of large-scale and high-quality datasets for training state-of-the-art vision models. \cite{zhai2106scaling} further explored scaling behavior in Vision Transformers~\citep{dosovitskiy2020image} using the  JFT-3B dataset, showing that scaling both data and model size simultaneously leads to improved generalization.

The pivotal role of data scaling is equally applicable to VLMs, as highlighted in \cite{cherti2023reproducible}. This has led to a substantial increase in the development of image-text datasets over the last ten years. Early datasets, such as COCO Captions~\citep{chen2015microsoft} and Flickr30k~\citep{young2014image}, were created to enable tasks like image captioning and visual question answering with high-quality annotations. However, their limited size, due to the cost of human annotation, hindered further scaling of the datasets. To address this, Conceptual Captions~\citep{sharma2018conceptual}  filtered image-text pairs from the web based on heuristic rules, leading to millions of image-caption pairs. Going forward along this road, larger image-text datasets have been created from web sources, using increasingly complex filtering techniques~\citep{datacomp,fang2023data,dong2025scalable}. These datasets, ranging from hundreds of millions to several billion image-text pairs, have enabled the training of powerful vision-language models like CLIP~\citep{radford2021learning} and ALIGN~\citep{jia2021scaling}. Notably, LAION-5B~\citep{schuhmann2022laion} and WebLI~\citep{chen2022pali} stand out as the largest publicly and privately available image-text datasets, with 5 billion and 10 billion multilingual image-text pairs respectively. However, the  growing web contains vastly more data. The impact of scaling to much larger datasets, such as 100 billion samples, is largely unknown.

\paragraph{Vision-Language Pre-training.}

The field of large vision-language models is advancing quickly, building upon remarkable progress in both vision and natural language processing. A prevalent and highly effective strategy is to learn visual representations and language modeling independently, followed by joint pre-training of the vision-language model using high-quality multimodal data.

Since the advent of CLIP~\citep{radford2021learning}, contrastive learning on large, noisy web datasets has become the dominant approach for acquiring powerful visual representations~\citep{chen2020simple,xu2024demystifyingclipdata}. This weakly supervised paradigm surpasses traditional supervised learning methods~\citep{kolesnikov2020big,steiner2021train}, primarily due to the large scale and high diversity of web data~\citep{jia2021scaling,yuan2021florence,pham2023combined,yu2022coca}. An alternative approach gaining traction involves learning visual features from web data using generative methods~\citep{tschannen2024image,wan2024locca}, which predict paired text for given images. While generative pretraining can provide superior transferability, the high computational cost limits its widespread adoption.

Despite the acquired zero-shot capabilities, which can be directly applied to tasks such as zero-shot classification~\citep{deng2009imagenet} and image-text retrieval~\citep{chen2015microsoft,young2014image}, the strong visual representations learned by contrastively trained models often lead to their utilization as image encoders. This is often leveraged in vision-language tasks by integrating visual tokens with language tokens, enabling LLMs to process multimodal information~\citep{alayrac2022flamingo,chen2022pali,chen2023pali,li2023blip,beyer2024paligemma,liu2024visual}. Following this approach, PaLI-3~\citep{chen2023pali} has demonstrated that vision models trained on large-scale web data outperform those trained on weakly annotated images of a similar scale, which further underscores the importance of the data diversity inherently present in the web corpus.

\paragraph{Inclusive Models.}
Recent studies have highlighted that popular techniques employed to enhance the performance of vision-language models, such as English-language-based filtering, may inadvertently diminish cultural understanding~\citep{goyal2022vision,nguyen2024multilingualdiversityimprovesvisionlanguage,richards2023does,pouget2024no,ananthram2024see}. In addition, quality filters, such as using CLIP similarity scores, were found to reduce the representation of demographic subgroups~\citep{hong2024s}.
This lack of representation is further exacerbated by the inherent long-tail distribution of web-scale datasets, leading to poor generalization on diverse visual categories~\citep{parashar2024neglected}.
Hence, we also evaluate cultural diversity, as outlined in~\cite{pouget2024no}, which falls into two categories.

The first category, geo-localization, involves predicting the country or region of origin for an image using few-shot classification.  The second category utilizes zero-shot classification on datasets curated from various geographical regions, such as Dollar Street~\citep{rojas2022dollar}, GeoDE~\citep{ramaswamy2024geode}, and Google Landmarks Dataset v2 (GLDv2)~\citep{weyand2020google}. These datasets include household items and landmarks from diverse geographic backgrounds, 
enabling the assessment of model performance on recognizing culturally important locations. In our evaluations, we employ all three datasets. For zero-shot evaluation on Dollar Street, we adhere to the methodology used in~\cite{rojas2022dollar}, mapping 96 specific topics within the dataset to corresponding ImageNet classes. This mapping results in a subset of 21K images.

\begin{table}[t]
    \centering\scriptsize
    \caption{\textbf{Error rates} (lower is better) on SigLIP-H/14. Scaling the data from 10B to 100B examples yields greater performance gains on culturally-relevant tasks than the traditional Western-centric tasks. See Appendix~\ref{appendix:scaling_law} for the full results.}
    \label{tab:main_mini}
    \begin{tabularx}{\columnwidth}{@{}l|YYY|YYY|YYY@{}}
    \toprule
    \bf Metric &\multicolumn{3}{c}{\textbf{Value}} &\multicolumn{6}{c}{\textbf{Scaling Laws}}\\
    &\multicolumn{3}{c}{\bf @100B ex} &\multicolumn{3}{c}{\bf exponent} &\multicolumn{3}{c}{\bf limit}\\
    &1B &10B &100B &1B &10B &100B &1B &10B &100B\\
    \midrule
    \multicolumn{10}{c}{\em Zero-shot classification}\\[5pt]

ImageNet 
&  29.6&25.6&\bf24.9&0.36&0.64&0.52&26.7&\underline{24.5}&\bf23.3
\\
CIFAR100 
& 23.5&\bf19.8&\underline{21.4}&0.25&0.36&0.29&20.6&\underline{18.0}&\bf17.6\\
Pet 
&10.3&\phantom{0}\underline{7.5}&\bf\phantom{0}7.2&0.45&0.42&0.50&\phantom{0}8.1&\phantom{0}\underline{5.3}&\bf\phantom{0}4.6\\\midrule
    \multicolumn{10}{c}{\em Retrieval @1}\\[5pt]

COCO I2T 
&  48.6&\bf42.0&\underline{42.5}&0.21&0.62&0.47&44.6&\underline{40.3}&\underline{40.6}\\
COCO T2I 
&  64.9&\underline{60.3}&\bf59.3&0.30&0.55&0.43&62.8&\underline{58.9}&\bf57.3\\
Flickr I2T
&  16.8&\bf13.5&\underline{13.9}&0.23&0.40&0.23&12.2&\underline{11.4}&\bf11.3\\
Flickr T2I 
& 34.3&\underline{28.5}&\bf28.0&0.23&0.56&0.46&29.6&\underline{26.8}&\bf25.9\\ \midrule

    \multicolumn{10}{c}{\em 10-shot}\\[5pt]

Imagenet 
&32.4&\underline{29.8}&\bf29.3&0.41&0.73&0.79&30.3&\underline{29.0}&\bf28.3\\

Birds 
&41.6&\underline{39.1}&\bf36.3&0.67&0.52&0.47&40.6&\underline{37.4}&\bf33.9\\

Caltech 
& \phantom{0}5.7&\phantom{0}\bf6.0&\phantom{0}\underline{8.9}&0.21&0.08&0.11&\phantom{0}4.3&\phantom{0}\bf3.7&\phantom{0}\underline{4.6}\\

Cars 
& 11.3&\underline{10.3}&\phantom{0}\bf9.6&0.27&0.88&0.44&\phantom{0}\underline{9.1}&10.1&\phantom{0}\bf8.3\\

CIFAR100 
& 25.8&\bf23.8&\underline{24.2}&0.22&0.25&0.24&21.4&\underline{21.1}&\underline{19.7}\\

Colorectal 
& \bf25.2&26.2&\underline{25.9}&0.22&0.20&0.15&\bf19.7&\underline{17.9}&{20.7}\\

Pet 
& 10.8&\phantom{0}\underline{9.1}&\phantom{0}\bf8.7&0.92&0.48&0.46&10.3&\phantom{0}\underline{7.6}&\phantom{0}\bf6.5\\

DTD 
& 29.2&\bf26.1&\underline{26.8}&0.16&0.23&0.23&25.0&\bf23.8&\underline{24.8}\\ \midrule

    \multicolumn{10}{c}{\em Cultural --- 10-shot Geolocalization}\\[5pt]

DS & 64.6&\underline{59.1}&\bf53.7&0.30&0.56&0.64&61.0&\underline{56.4}&\bf52.5\\
GeoDE-C 
&56.9&\underline{50.2}&\bf47.6&0.23&0.78&0.62&52.2&\underline{49.4}&\bf46.1\\
GeoDE-R  
&54.6&\underline{47.6}&\bf44.7&0.00&0.38&0.31&50.1&\underline{45.3}&\bf41.0\\\midrule

    \multicolumn{10}{c}{\em Cultural --- Zero-shot classification}\\[5pt]

DS &50.0&\underline{48.6}&\bf47.4&0.15&0.13&0.20&\bf43.9&44.2&\underline{44.1}\\
GeoDE
&  \phantom{0}6.0&\phantom{0}\underline{4.9}&\phantom{0}\bf4.8&0.19&0.22&0.24&\phantom{0}\bf3.3&\phantom{0}\bf3.3&\phantom{0}\underline{3.5}\\
GLDv2 
&48.1&\underline{40.1}&\bf38.8&0.52&1.34&0.80&46.0&\underline{39.0}&\bf36.8\\ \bottomrule
    \end{tabularx}
\end{table}

\section{Experimental Setup}\label{sect:setup}

\subsection{Pre-training Datasets}

\paragraph{Raw Datasets.}To assess the performance of vision-language models on large-scale image-text data, we construct a dataset with 100 billion image-text pairs from the web, inspired by the work of \cite{chen2022pali,schuhmann2022laion,zhai2022lit,jia2021scaling}. We refer to this as WebLI-100B, and refer to its subsets with 1 billion and 10 billion examples as 1B and 10B, respectively.  The 1B and 10B datasets are created by randomly sampling 1\% and 10\%, respectively, from the 100 billion dataset. We employ a minimal set of essential data filters: removing harmful images with the Google Cloud Vision API (SafeSearch Detection) and personally identifiable information (PII) using the Google Cloud Data Loss Prevention (DLP) API. We do not apply any quality or language filtering, which ensures the dataset remains as multilingual and diverse as possible. We utilize both the alt-text and page title associated with each image as the paired text. To ensure fair evaluations, we de-duplicate images across +90 common vision-language tasks, following the methodology described  in~\cite{chen2022pali}. We use the multilingual tokenizer mC4~\cite{xue2021mt5massivelymultilingualpretrained}.

\paragraph{Quality-filtered Datasets.}To examine the impact of scaling on quality-filtered data, we adopt the common approach of using the CLIP-L/14 model~\citep{radford2021learning} as a filter, retaining a high-quality dataset with 5 billion pairs of images and English alt-text. To further solidify our results, we train a VLM on the web data to classify image-text pairs as aligned or misaligned, and tune its threshold to retrain another filtered dataset of the same size. Unless otherwise noted, we use the language of web pages\footnote{The ``content-language" meta tag in the head of an HTML document.} for multilingual experiments to avoid potential errors in language detection on noisy web text.

\paragraph{Language-rebalanced Datasets.}In the language rebalancing experiments in Section~\ref{sec:lang_rebalance}, we adjust the mixing ratio of the low-resource languages used in  Crossmodal-3600~\citep{thapliyal2022crossmodal}. These low-resource languages are Bengali (bn), Filipino (fil), Hindi (hi), Hebrew (iw), Maori (mi), Swahili (sw), and Telugu (te)\footnote{Cusco Quechua (quz) is excluded from our experiments because it is not supported by our language detection method.}, ranging from 0.001\% to 0.267\% in our dataset (Appendix~\ref{appendix:lang_distribution}). In model training, we upsample each of them to 1\%, with remaining 93\% comprising of the original data.

\subsection{Contrastive Vision-Language Pre-Training}
To study the impact of data scale on model performance, we train SigLIP~\citep{zhai2023sigmoidlosslanguageimage} models using the three different dataset sizes: 1 billion, 10 billion and 100 billion. 
We also vary the model size using ViT-B/16, ViT-L/16, and ViT-H/14 architectures for both image and text encoders. During contrastive training, inspired by \cite{zhai2106scaling}, we utilize a large batch size of 32K and an inverse square root learning rate schedule with 200 million warmup and cooldown examples. The learning rate and weight decay are set to 0.001 and 0.0001 respectively. In the preprocessing stage, images are resized to a resolution of 224x224 pixels, and texts are tokenized using the multilingual mt5~\citep{xue2020mt5} tokenizer with a maximum sequence length of 64 tokens. Models are trained on $16\times16$ TPUv5 chips. 

All models are trained on a maximum of 100 billion examples with 0.5 token dropping; e.g. a maximum of 100 epochs when using 1B examples. We cool down the models at various training steps where they have seen 3, 7, 10, 17, 26, 33, 49, 66, and 100 billion examples, and evaluate them after the cool-downs. Unless otherwise specified, we report results using the checkpoints where models have been trained on 100 billion examples. All models are compared on a compute-matched regime.

\subsection{Evaluations}\label{sect:evals}
The model's capabilities are evaluated across a diverse range of benchmarks, spanning from traditional Western-centric tasks to those measuring inclusivity.

\paragraph{Western-centric.} Our first set of evaluations uses diverse, well-established benchmarks. For zero-shot classification, we employ ImageNet~\citep{deng2009imagenet}, CIFAR-100~\citep{krizhevsky2009learning}, and Oxford-IIIT Pet~\citep{pets} datasets. Additionally, for 10-shot evaluations, we use Caltech-UCSD Birds~\citep{wah2011caltech}, Caltech 101~\citep{li_andreeto_ranzato_perona_2022}, Cars196
~\citep{KrauseStarkDengFei-Fei_3DRR2013}, Colorectal Histology~\citep{kather2016multi}, and Describable Textures Dataset (DTD)~\citep{cimpoi14describing} benchmarks to assess the representation capabilities of vision models. We also conduct zero-shot retrieval evaluations on COCO Captions~\citep{chen2015microsoft} and Flickr30k~\citep{young2014image}, in both image-to-text and text-to-image directions.

\paragraph{Cultural Diversity.}
Besides the above metrics, we also incorporate a range of benchmarks aimed at evaluating cultural diversity, following the recommendations in~\cite{pouget2024no}. Specifically, we include zero-shot classification using Dollar Street~\citep{rojas2022dollar}, GeoDE~\citep{ramaswamy2024geode}, and Google Landmarks Dataset v2 (GLDv2)~\citep{weyand2020google}. See Section~\ref{sect:related_work} for a brief description about each dataset. We also use 10-shot geolocalization using Dollar Street and GeoDE. 

\paragraph{Multilinguality.} We evaluate the model's multilinguality using the Crossmodal-3600 dataset~\citep{thapliyal2022crossmodal}, a geographically diverse set of 3600 images with human-generated captions in 36 languages. We assess the model's zero-shot retrieval in both image-to-text and text-to-image directions for each language. In addition to per-language results, we also present average scores for low-resource languages (Bengali, Filipino, Hindi, Hebrew, Maori, Swahili, and Telugu) and high-resource languages (others).

\paragraph{Fairness.}
In addition, we also evaluate the presence of societal biases in the trained model. We report on representation bias (RB) and association bias (AB) between gender and occupation, as defined in~\cite{alabdulmohsin2024clip}. These measure unwanted associations w.r.t. the gender attribute using 1st and 2nd order statistics. Also, we report performance disparity by income in Dollar Street zero-shot accuracy and by region in GeoDE zero-shot accuracy.

\paragraph{Transfer to Generative Models.}
Finally, to assess how well our contrastively trained vision models transfer to generative vision-language tasks, we utilize the PaliGemma model~\citep{beyer2024paligemma}. We initialize PaliGemma's vision component with our contrastively trained models and pretrain it on 50 million seen examples, following its stage-1 recipe. During pretraining, we explore two common transfer settings: freezing~\citep{liu2024visual,zhu2023minigpt,chen2022pali} and unfreezing~\citep{xiao2024florence,chen2023pali,beyer2024paligemma,steiner2024paligemma2familyversatile} the vision model. We then finetune on a variety of downstream tasks using the default configuration.
\section{Experimental Results}\label{sect:results}

\subsection{Established Benchmarks}\label{sect:results_west}
We begin by evaluating all VLMs on established benchmarks, based on ImageNet and COCO Captions, among other datasets. Results for  SigLIP-H/14 are provided in Table~\ref{tab:main_mini}, whereas full results are in Appendix~\ref{appendix:scaling_law}.  Our analysis reveals that increasing the dataset size from 10 billion to 100 billion examples does not improve performance substantially. This is statistically supported by Wilcoxon's signed rank test~\citep{wilcoxon1992individual}, which gives a $p$-value of 0.9, indicating that differences are not significant.

In addition, we also fit data scaling laws for every combination of model and dataset following the recipe proposed in~\cite{alabdulmohsin2022revisiting}. This allows us to evaluate whether or not the performance gap is expected to increase or decrease in the infinite-compute regime. We report the resulting scaling exponents and asymptotic performance limits in the tables. Again, we do not observe  significant differences at the 95\% confidence level ($p$-value of 0.09).

\subsection{Cultural Diversity}
Unlike the Western-oriented metrics reported in Section~\ref{sect:results_west}, cultural diversity metrics present an entirely different picture. We observe \emph{significant} gains when scaling the size of the dataset from 10 billion to 100 billion examples. For SigLIP-H/14, this is shown in Table~\ref{tab:main_mini}, and the full results are provided in Appendix~\ref{appendix:scaling_law}. Using Wilcoxon's signed rank test, we obtain a $p$-value of 0.002, indicating a statistically significant evidence at the 99\% confidence level.

\subsection{Multilinguality}
\begin{figure}[h]
    \includegraphics[width=\linewidth]{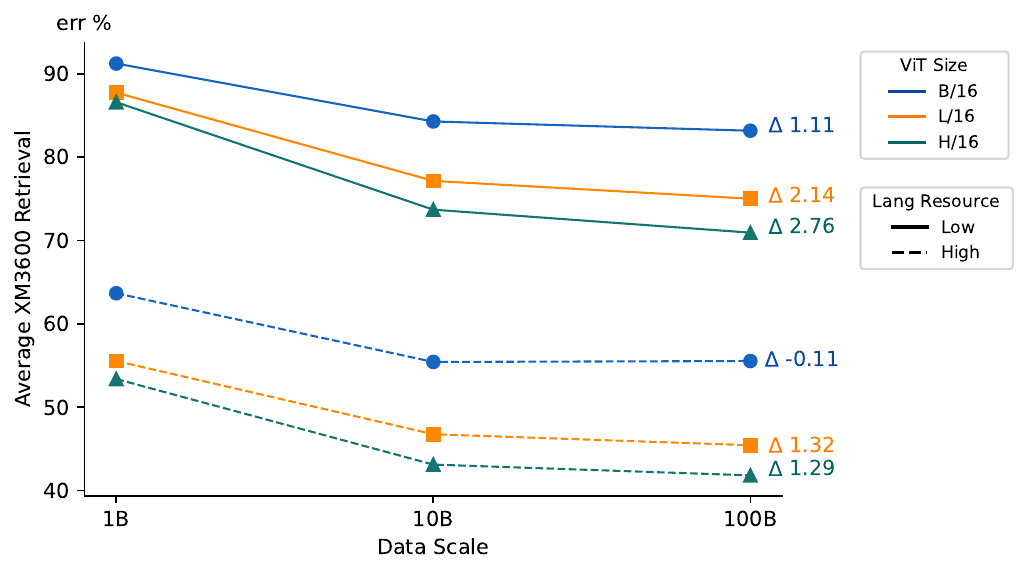}
    \caption{Scaling up to 100B examples leads to more notable improvements in low-resource languages. $\Delta$ denotes the improved accuracy when scaling from 10B examples to 100B.}
    \label{fig:multilinguality}
\end{figure}

Our multilingual benchmark, Crossmodal-3600 zero-shot retrieval~\citep{thapliyal2022crossmodal}, shows a disparity in performance gains: low-resource languages benefit more from the 100 billion scale than high-resource ones. The disparity, illustrated in Figure~\ref{fig:multilinguality}, which not only exists in all model sizes but also widens as the models become larger. Detailed results for each language are in Appendix~\ref{appendix:data_scale}. Later in Section~\ref{sec:lang_rebalance}, we investigate the impact of language balancing.

\begin{table}[t!]
\centering
\footnotesize

\begin{tabular}{p{0pt}l|rrrrr}
\toprule
& Data & Semantics & OCR & Multiling & RS & Avg \\
\midrule
\includegraphics[width=8pt]{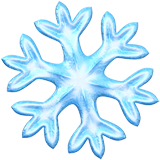} & 1B & 76.0 & 66.8 & 67.0 & 92.3 & 73.6 \\
\includegraphics[width=8pt]{images/snowflake_2744-fe0f.png} & 10B & 75.4 & 65.2 & 66.3 & 91.9 & 72.7 \\
\includegraphics[width=8pt]{images/snowflake_2744-fe0f.png} & 100B & 76.4 & 67.0 & 66.9 & 92.1 & 73.9 \\
\includegraphics[width=8pt]{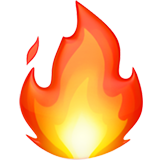} & 1B & 77.1 & 69.5 & 66.9 & 92.0 & 75.1 \\
\includegraphics[width=8pt]{images/fire_1f525.png} & 10B & 76.4 & 66.9 & 66.0 & 91.8 & 73.7 \\
\includegraphics[width=8pt]{images/fire_1f525.png} & 100B & {77.2} & {70.0} & {67.0} & {91.8} & {75.3} \\
\bottomrule
\end{tabular}

\caption{The PaliGemma transfer results of ViT-L/16 models pretrained on 10B and 100B examples, with both frozen (\includegraphics[width=8pt]{images/snowflake_2744-fe0f.png}) and unfrozen (\includegraphics[width=8pt]{images/fire_1f525.png}) vision components. Results are aggregated by task.}
\label{tab:transfer_avg}
\end{table}

\subsection{Fairness}
We report the three metrics discussed in Section~\ref{sect:evals}. 

\paragraph{Representation Bias.} The first metric is representation bias (RB), with results in Table~\ref{tab:rb}. We note that models trained on raw web data (without rebalancing) have a significantly higher preference to associate a randomly chosen image from ImageNet~\citep{deng2009imagenet} with the label ``Male'' over the label ``Female.''

\noindent
\begin{minipage}{0.5\linewidth}
In fact, this occurs nearly 85\% of the time. Training on 100B examples does not mitigate this effect. This finding aligns with previous research highlighting the necessity of bias mitigation strategies, such as data balancing~\cite{alabdulmohsin2024clip}, to address inherent biases in web-scale datasets.
\end{minipage}
\hfill
\begin{minipage}{0.46\linewidth}
\centering\scriptsize

\begin{tabularx}{\columnwidth}{@{}l|YYY@{}}
    \toprule
    \bf Model&\bf1B &\bf10B &\bf100B\\
    \midrule
B & 83.2&84.5&85.2\\
L & 88.2&86.4&85.5\\
H & 86.8&85.0&86.6\\
\bottomrule
\end{tabularx}
\captionof{table}{Representation bias w.r.t. gender (see Section~\ref{sect:results}). Here,  values [\%] indicate how often model prefers to associate a random  image with label ``Male'' over ``Female''.} \label{tab:rb}
\end{minipage}

\begin{figure}[h]
    \centering
    \includegraphics[width=0.495\columnwidth]{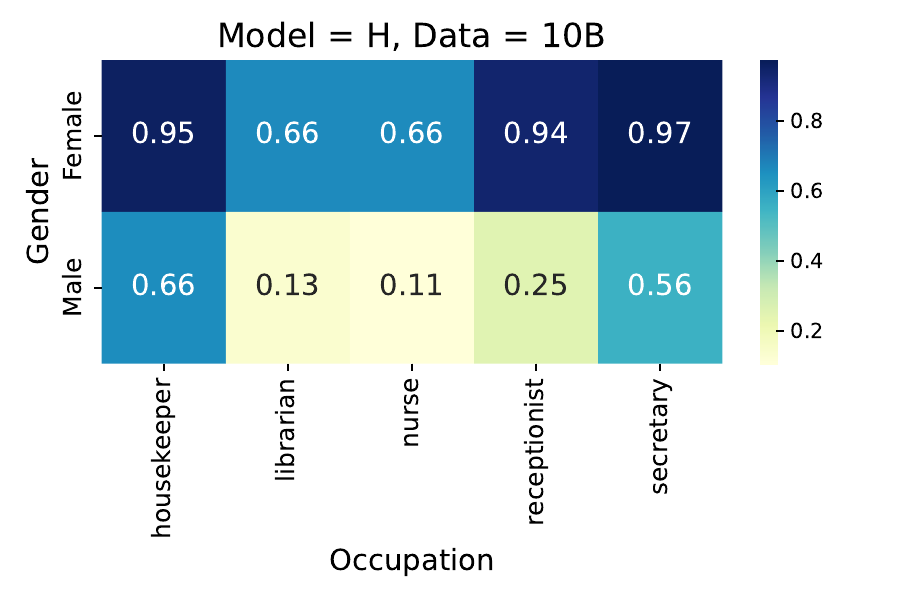}
    \includegraphics[width=0.495\columnwidth]{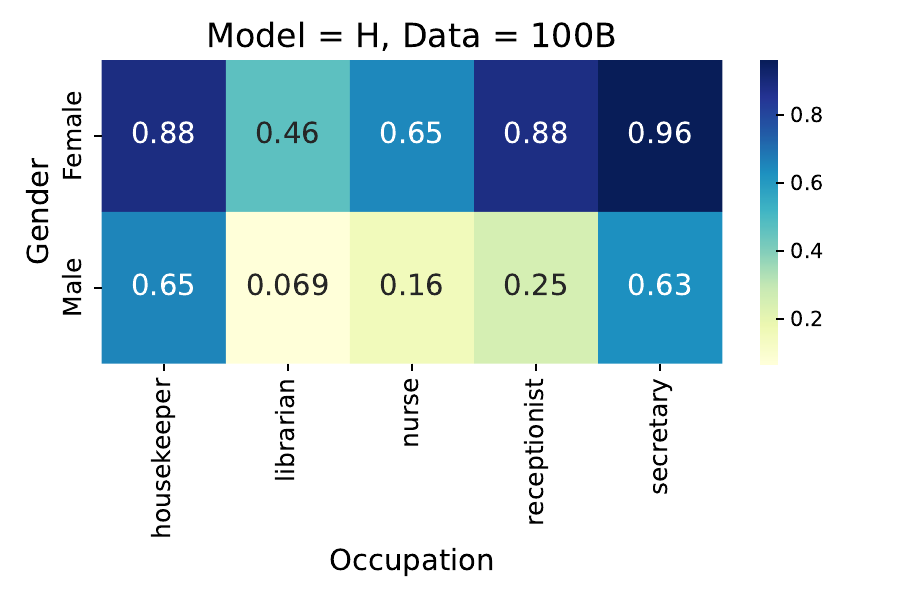}
    \caption{Association bias between gender and occupation in SigLIP-H/14 pretrained on 10B (left) and 100B (right) examples. Increasing the data scale does not mitigate such biases.}
    \label{fig:ab}
\end{figure}

\paragraph{Association Bias.} Second, Figure~\ref{fig:ab} shows the association bias in SigLIP-H/14 between gender and occupation as we scale the data from 10B to 100B examples. Specifically, we plot the probability that the model prefers a particular occupation, such as ``{\fontfamily{lmodern}\selectfont secretary}'' over another, such as ``{\fontfamily{lmodern}\selectfont manager}'' when images correspond to males or females. In this evaluation, we use the Fairface~\citep{karkkainen2021fairface} dataset. The labels we compare are: ``{\fontfamily{lmodern}\selectfont librarian}'' vs. ``{\fontfamily{lmodern}\selectfont scientist}'', ``{\fontfamily{lmodern}\selectfont nurse}'' vs. ``{\fontfamily{lmodern}\selectfont doctor}'', ``{\fontfamily{lmodern}\selectfont housekeeper}'' vs. ``{\fontfamily{lmodern}\selectfont homeowner}'', ``{\fontfamily{lmodern}\selectfont receptionist}'' vs. ``{\fontfamily{lmodern}\selectfont executive}'' and ``{\fontfamily{lmodern}\selectfont secretary}'' vs. ``{\fontfamily{lmodern}\selectfont manager}''. Again, we do not see a reduction in association bias by scaling up the training data. Full results are in Appendix~\ref{appendix:ab}.



\paragraph{Performance Disparity.} Finally, one common definition of fairness in machine learning is maintaining similar performance across different groups. See, for instance,~\cite{dehghani2023scaling} and the related notions of ``Equality of Opportunity'' and ``Equalized Odds''~\citep{hardt2016equalityopportunitysupervisedlearning}.

\noindent
\begin{minipage}{0.44\linewidth}
To evaluate this, we calculate disparity (lower is better) across income level in Dollar Street and across geographic region in GeoDE. Table~\ref{tab:per_disp_mini} and the full results in Appendix~\ref{appendix:perf_disp} show that scaling the data to 100 billion examples \emph{improves} performance disparity.
\end{minipage}
\hfill
\begin{minipage}{0.52\linewidth}
\centering\scriptsize
    \begin{tabularx}{\columnwidth}{l|YYY@{}}
    \toprule
    \bf Model & \bf 1B & \bf 10B & \bf 100B\\ \midrule
    &\multicolumn{3}{c}{\em 0-shot Dollar Street}\\[2pt]
B & 32.5 & 29.9 & \bf29.0\\
L & \bf29.7 & 29.8 & 30.4 \\
H & 32.2 & 33.0 & \bf32.1\\
\midrule 
    &\multicolumn{3}{c}{\em 0-shot GeoDE}\\[2pt]
B & 4.7 & 5.5 & \bf4.4\\
L & 3.2 & 4.0 & \bf2.8 \\
H & 3.6 & 3.0 & \bf2.7\\
\bottomrule
    \end{tabularx}
    \vspace{-7pt}
    \captionof{table}{Performance disparity for models pretrained on 100B seen examples of different data scales.}
    \label{tab:per_disp_mini}
\end{minipage}


\subsection{Transfer To Generative Models}
\label{sec:transfer}

We use PaliGemma~\citep{beyer2024paligemma} with both frozen and unfrozen vision component to assess the transferability of our vision models, which were contrastively pre-trained on datasets of different scales. Here, we pre-train with stage 1 recipe, and pre-train PaliGemma on 50 million seen examples. In Table~\ref{tab:transfer_avg}, when taking the noise level into account (such as the unexpected performance drop when scaling data from 1B to 10B), we do not observe consistent performance gains across downstream tasks as we scale the pretraining dataset. See Appendix~\ref{appendix:transfer} for full details.


\section{Analysis}\label{sect:analysis}

\subsection{Data Quality Filtering}
\label{sec:data_filter}

\begin{figure*}[t]
    \includegraphics[width=.33\linewidth]{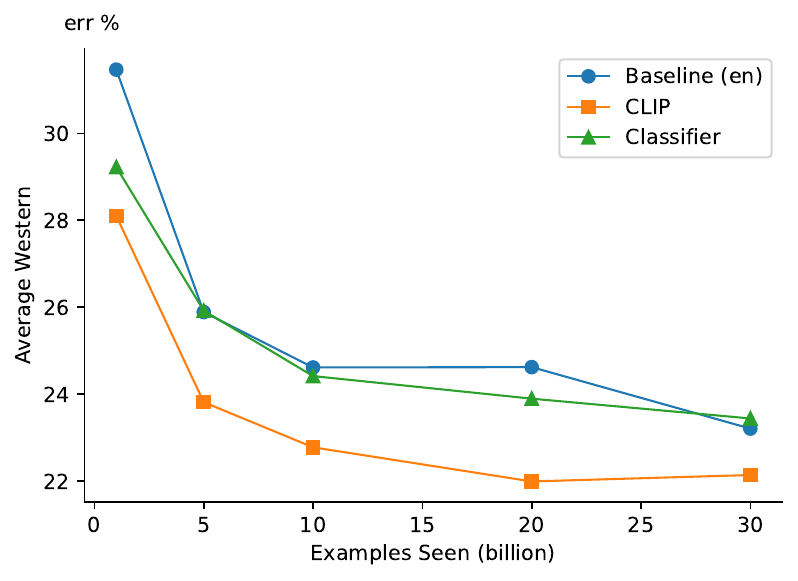}
    \includegraphics[width=.33\linewidth]{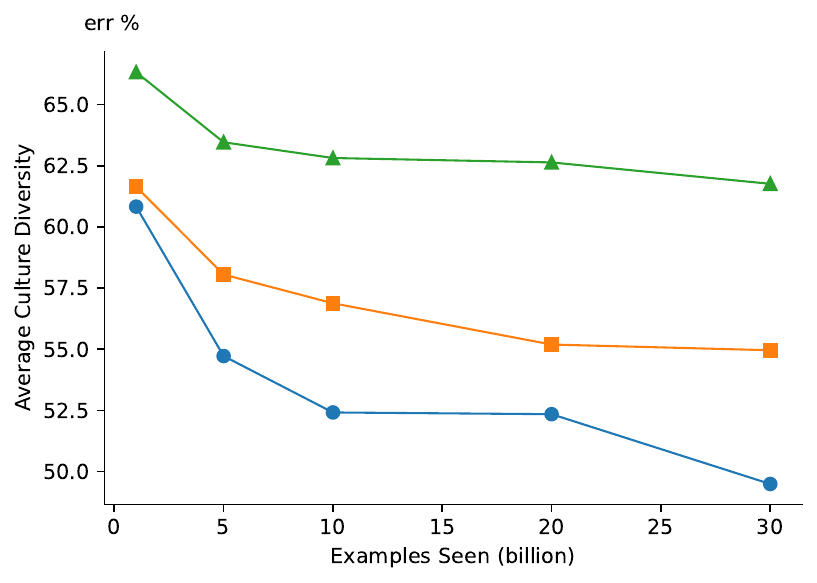}
    \includegraphics[width=.33\linewidth]{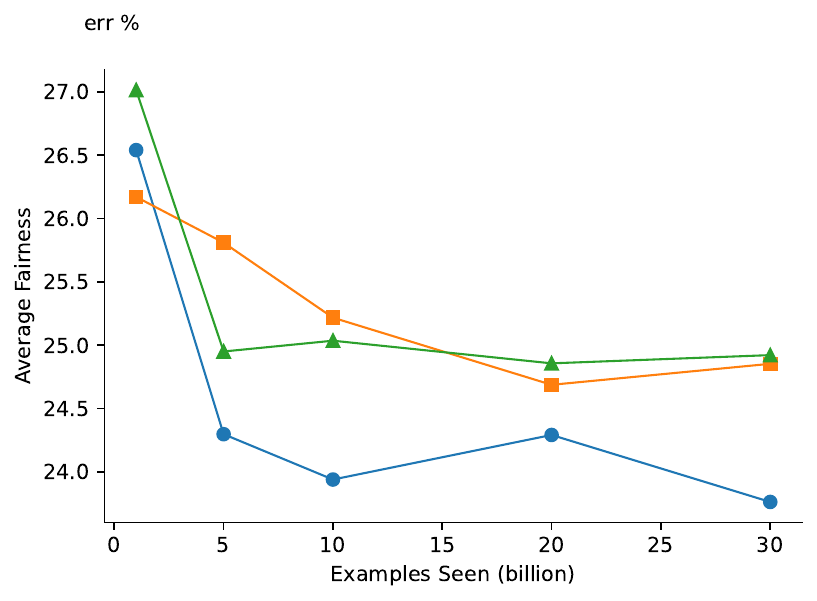}
    \caption{Filters can hurt  diversity (middle) and fairness (right), while benefiting Western-centric tasks (left). This holds both when filtering by CLIP or a classifier trained on multilingual data.}
    \label{fig:quality_filter}
\end{figure*}

Raw web data can be noisy for training effective vision-language models. To address this, a common strategy is to use a data filter model to remove less relevant image-text pairs. In this work, we utilize the CLIP-L/14 model to filter the raw data and retrain 5 billion high-quality English image-text pairs.
For comparison, we also train a classifier model on the raw web data, resulting in a filtered dataset of the same size. Additionally, we sample an English subset of the same size from the raw data as a baseline.
We train ViT-L models on the three datasets and present the results in Figure~\ref{fig:quality_filter} and Appendix~\ref{appendix:quality_filter}. CLIP filter excels in Western-centric tasks in agreement with prior work~\citep{fang2023data,cao2023less,maini2023t,abbas2023semdedup,goyal2024scalinglawsdatafiltering}. However, all filtered datasets underperform in other tasks, particularly those involving cultural diversity. This illustrates a key drawback of data filtering: it can inadvertently introduce biases into the filtered dataset, in agreement with prior works~\citep{birhane2021multimodal,pouget2024no,garcia2023uncurated}.

\subsection{Language Rebalancing}
\label{sec:lang_rebalance}

\begin{minipage}{0.48\linewidth}
As shown in Table~\ref{tab:lang_distribution_mini}, low-resource languages collectively represent only 0.8\% in our raw data, which prevents sufficient model learning of the concepts existing in these languages or areas. To address this imbalance, we upsample each low-resource language to a fixed 1\% representation. 
\end{minipage}
\hfill
\begin{minipage}{0.48\linewidth}
\centering\scriptsize

\begin{tabularx}{\linewidth}{l|Y}
\toprule
Language & (\%) \\
\midrule

\scriptsize Maori &  \scriptsize 0.001 \\
\scriptsize Telugu &\scriptsize  0.036 \\
\scriptsize Swahili & \scriptsize 0.046 \\
\scriptsize Filipino & \scriptsize 0.111 \\
\scriptsize Bengali & \scriptsize 0.113 \\
\scriptsize Hebrew & \scriptsize 0.240 \\
\scriptsize Hindi & \scriptsize 0.267 \\
$\ldots$ & $\ldots$\\
\scriptsize Japanese & \scriptsize 8.752 \\
\scriptsize English &\scriptsize  35.353 \\
\scriptsize\textbf{Low-resource All}  &\scriptsize 0.814 \\
\scriptsize\textbf{High-resource All}  &\scriptsize 94.510 \\\bottomrule
\end{tabularx}
\vspace{-7pt}
\captionof{table}{Language Distribution.}
\label{tab:lang_distribution_mini}
\end{minipage}

\begin{figure}
    \centering
    \includegraphics[width=0.495\columnwidth]{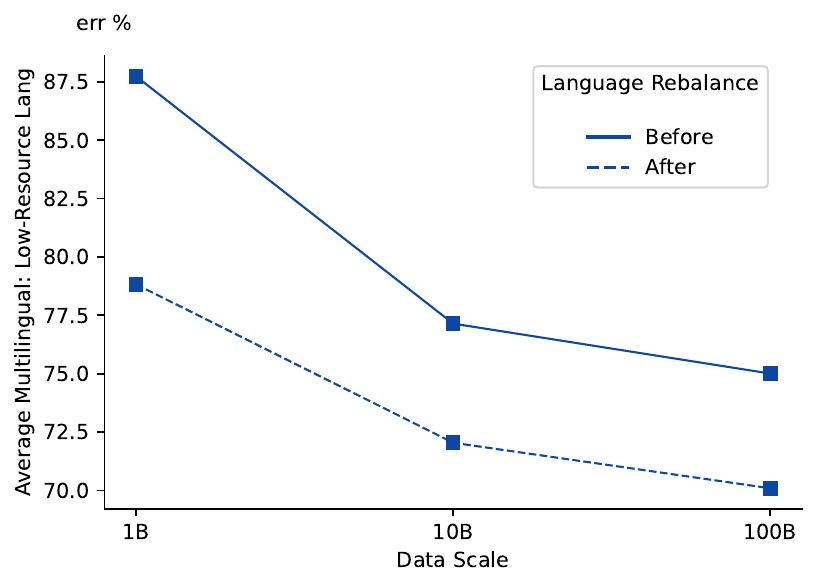}
    \includegraphics[width=0.495\columnwidth]{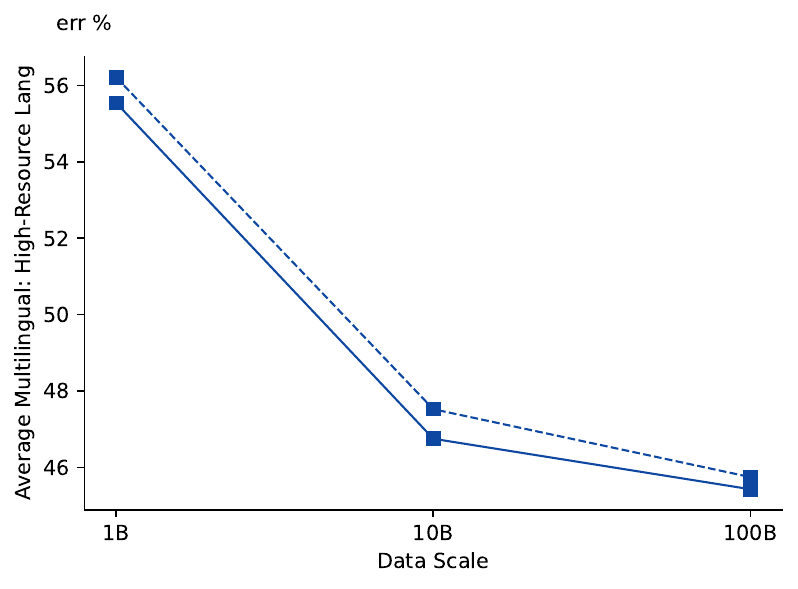}
    \caption{Impact of language rebalancing. Rebalancing significantly boosts performance on low-resource languages while maintaining on-par performance for high-resource languages.}
    \label{fig:lang_relance_mini}
\end{figure}

This rebalancing, visualized in Figure~\ref{fig:lang_relance_mini}, improves model performance on the low-resource language benchmark. 
While performance on the high-resource language slightly decreases, it remains comparable across relevant English-only tasks. This results in an overall improvement on the entire multilingual benchmark, consistent with findings in \cite{chuang2025meta}.
Full  results are in Appendix~\ref{appendix:lang_rebalance}.

\subsection{Qualitative Examples}
\label{sec:qualitative_examples}

We visualize the attention maps from the vision models trained on different scales of data in Table~\ref{tab:attention_maps_mini}. Models trained on larger data tend to have more focused attention on semantically relevant regions.
For example, in the ``Igloo'' image, the 100B-trained model accurately focuses on the igloo' structural details as well as the presence of icy mountains in the background.
Beyond low-resource concepts, 100B data can also improve performance on common concepts. As shown in the ``Bison" image, models trained on larger datasets more precisely capture the bison, rather than the surrounding landscape.
More visualized examples can be found in Table~\ref{tab:attention_maps}.


\begin{table*}[!t]
\centering
\footnotesize
\caption{The attention map visualization of the ViT-L/16 models trained on different scales of data. Images are selected to represent cultures in Western-centric countries and countries where low-resource languages are spoken.}
\label{tab:attention_maps_mini}

\begin{tabularx}{\textwidth}{@{}l@{} >{\centering\arraybackslash}X@{} >{\centering\arraybackslash}X@{} >{\centering\arraybackslash}X@{} >{\centering\arraybackslash}X @{}}
\toprule
\textbf{Concept} & \textbf{Image} & \textbf{1B Data} & \textbf{10B Data} & \textbf{100B Data} \\

Igorot Dance (Igorot)
&
\includegraphics[width=0.6\linewidth,height=0.6\linewidth]{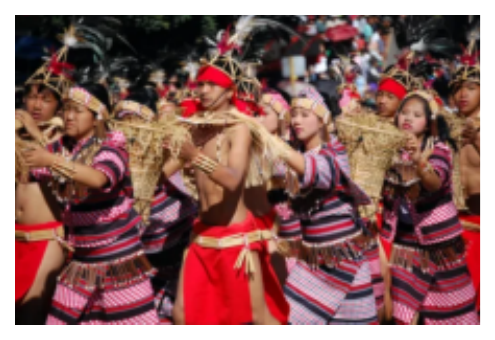} &
\includegraphics[width=0.6\linewidth]{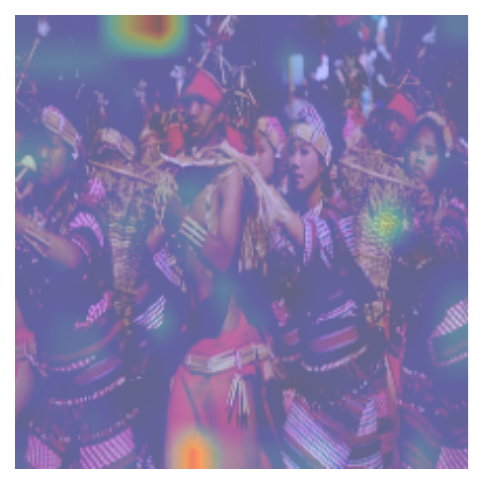} &
\includegraphics[width=0.6\linewidth]{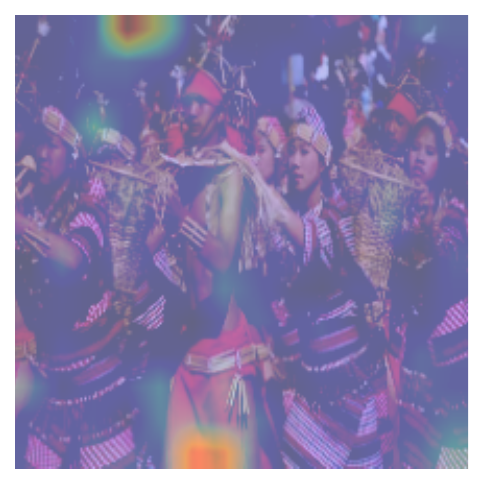} &
\includegraphics[width=0.6\linewidth]{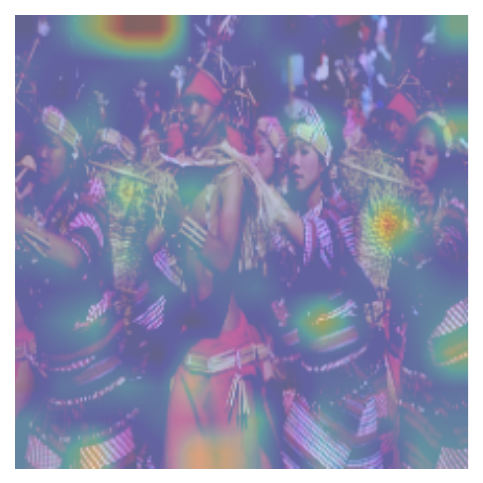} \\

Igloo (Inuit)
&
\includegraphics[width=0.6\linewidth,height=0.6\linewidth]{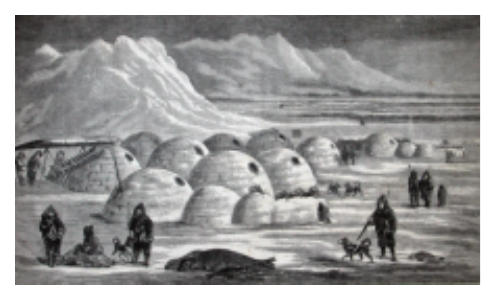} &
\includegraphics[width=0.6\linewidth]{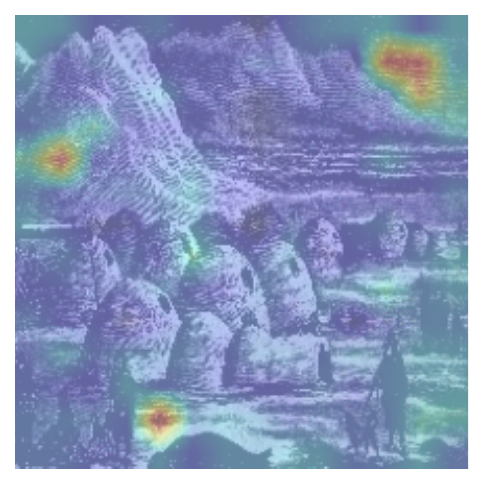} &
\includegraphics[width=0.6\linewidth]{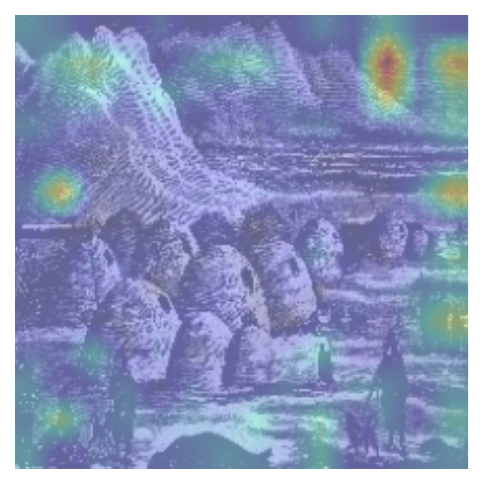} &
\includegraphics[width=0.6\linewidth]{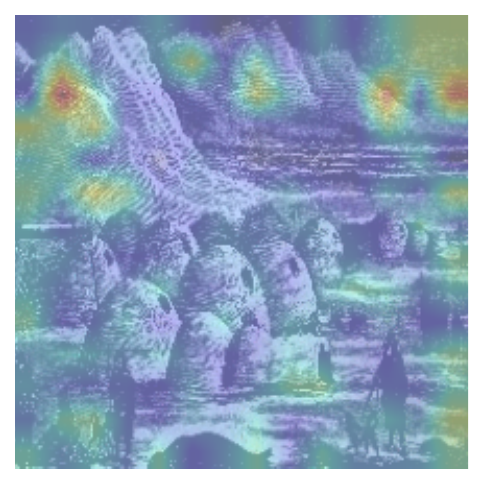} \\

\hline
Bison (Yellowstone)
&
\includegraphics[width=0.6\linewidth,height=0.6\linewidth]{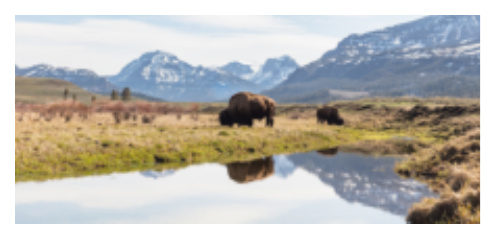} &
\includegraphics[width=0.6\linewidth]{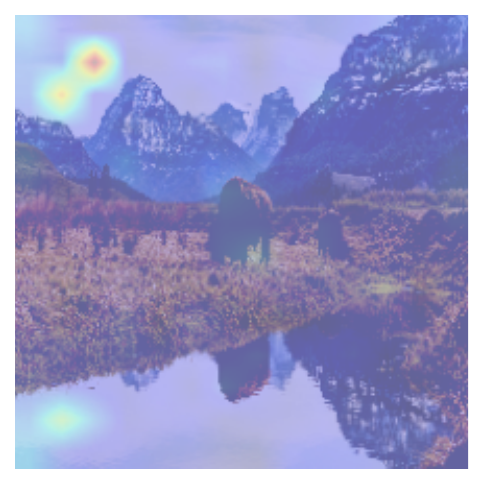} &
\includegraphics[width=0.6\linewidth]{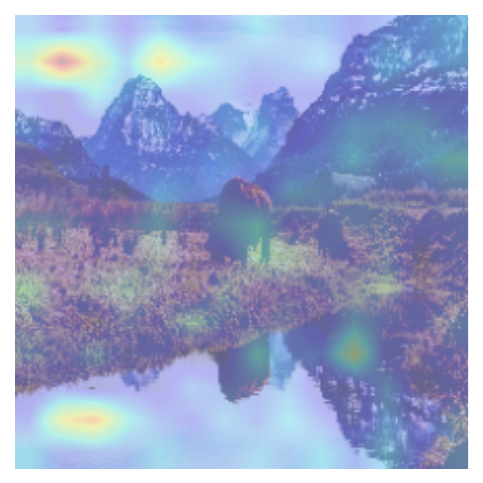} &
\includegraphics[width=0.6\linewidth]{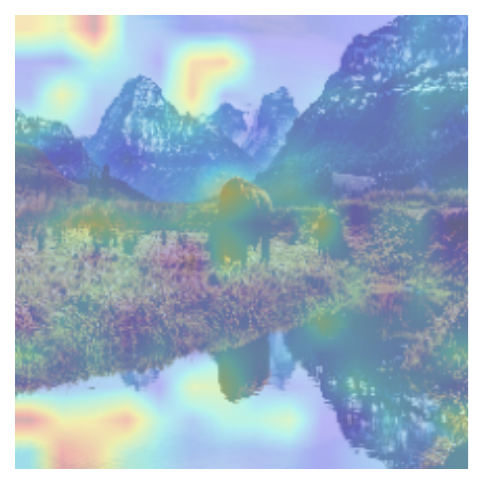} \\[5pt]

\bottomrule
\end{tabularx}
\end{table*}

\subsection{Scaling Law}
\label{sec:compute_scaling_law}

\begin{figure}[t]
\begin{subfigure}{\textwidth}
    \includegraphics[width=0.24\columnwidth]{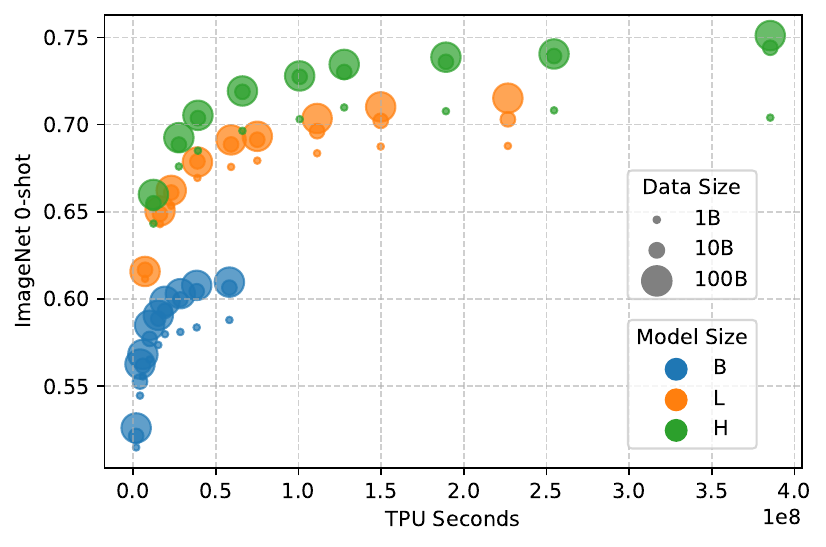}
    \includegraphics[width=0.24\columnwidth]{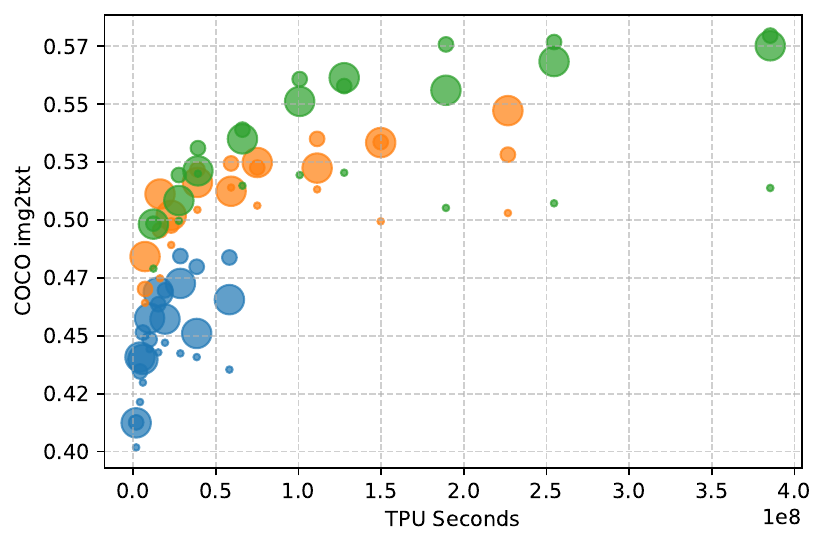}
    \newline
    \includegraphics[width=0.24\columnwidth]{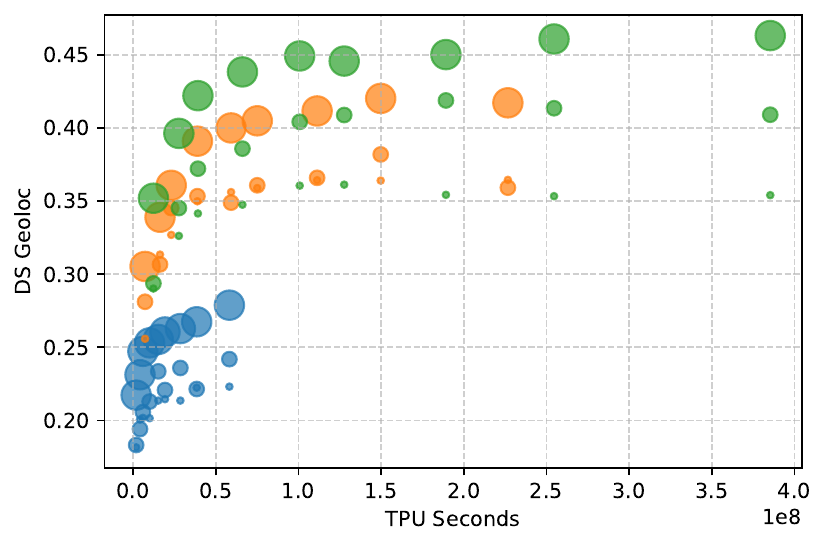}
    \includegraphics[width=0.24\columnwidth]{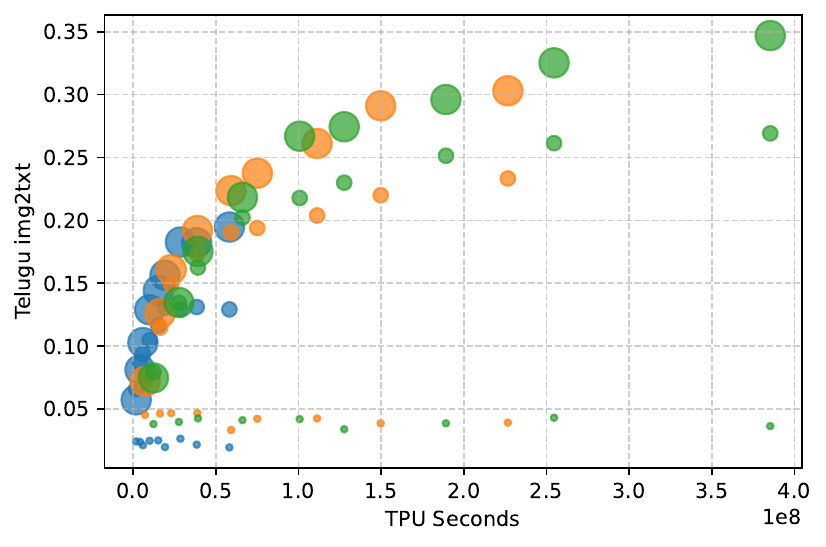}
\end{subfigure}
    \caption{Impact of scaling data, model and compute.}\vspace{-10pt}
    \label{fig:compute_scaling_law}
\end{figure}

As shown in Figure~\ref{fig:compute_scaling_law}, performance gains for Western-centric tasks (ImageNet, COCO) are marginal when scaling the dataset from 10 to 100 billion unique examples. However, tasks involving cultural diversity (DS Geoloc) and low-resource languages (Telugu) continue to improve with data size. Compute scaling benefits all tasks, suggesting the value of continued training with non-unique data, consistent with OpenCLIP observations~\cite{cherti2023reproducible}.

\subsection{Absolute Performance Levels}
\label{sec:absolute_perf}

Varying absolute performance levels across downstream tasks can confound claims of saturation or improvement. Saturation often correlates with high-performance regimes, whereas notable gains are often observed in low-performance regimes.
To investigate this, we gather data for the H model on the Crossmodal-3600 benchmark~\cite{thapliyal2022crossmodal}, which spans a wide performance spectrum (Appendix~\ref{appendix:absolute_perf}). This analysis confirms a weak correlation between absolute performance and performance gains.
\section{Discussion}\label{sect:discuss}

\paragraph{Reproducibility and Data Release.} We are committed to maximizing reproducibility for the vision-language community. Due to standard industrial policy regarding proprietary assets, we are unable to release our raw data, metadata files, or pipeline code. For maximal transparency and to facilitate dataset reconstruction, our dataset is derived exclusively from publicly available web pages, without applying quality or language filters.
Consequently, any researcher can replicate our image-text pairs by extracting the corresponding images, alt-texts, and page titles from the public web as in \cite{schuhmann2022laion,xu2024demystifyingclipdata,jia2021scaling}. This process can be efficiently executed using open-source tools such as MetaCLIP~\cite{xu2024demystifyingclipdata} and img2dataset~\cite{beaumont2021img2dataset} pipelines. Furthermore, the Common Crawl archive, which contains more than 250 billion pages, represents an even larger potential data source than the one we utilized.

\paragraph{Comparison with OpenCLIP.} Our ImageNet results are lower than those reported in OpenCLIP due to several factors. First, we pretrain on \emph{multilingual} data, whereas OpenCLIP uses English only, which is known to enhance performance on Western-oriented metrics like ImageNet~\cite{pouget2024no}. Second, we use 50\% token dropping to decrease computational overhead. Third, we apply minimal data filtering to eliminate its confounding effects, thereby focusing our research on the impact of scaling unfiltered data. 



\paragraph{Limitations.} The benchmarks used to evaluate VLM inclusivity are necessarily limited because inclusivity, a broad societal concept, cannot be reduced to a few metrics. For example, our use of Crossmodal-3600 for 0-shot multilinguality  is constrained by its coverage of only 36 languages and 3,600 images. Additionally, model sizes were restricted to ViT-H (~600M parameters) due to the large data scale. However, we do not consider model size a bottleneck; prior work~\cite{hooker2020characterisingbiascompressedmodels} suggests that if it were, the impact on long-tail cultural concepts would be significant, and our setup shows substantial cultural improvements with data scaling, indicating model size is not a limiting factor. Also, while we conduct data scaling laws, joint scaling laws for both model and compute—which would provide a more complete picture—are computationally prohibitive given our data scales.




\paragraph{Environmental Impact.} Training on 100 billion examples demands substantial computational resources. We minimize the environmental impact of this scale by running all of our experiments in Google Cloud, which is powered by 100\% renewable energy.

\section{Conclusion}








In this paper, we investigate the impact of scaling image-text data up to 100 billion unique examples, on vision-language pre-training. We demonstrate that a scale of 100 billion image-text pairs is beneficial for VLMs in areas beyond traditional Western-centric benchmarks, such as cultural diversity, multilinguality, and reducing performance disparity. Hence, this data scale remains important for the development of truly inclusive multimodal systems. We also investigate the impact of applying quality filters to large-scale image-text datasets. These filters, though often beneficial for traditional tasks, can negatively impact data diversity by reducing the representation of certain cultural contexts. Finally, multilinguality also benefits from the 100B data scale, particularly for low-resource languages.

{
    \small
    \bibliographystyle{ieeenat_fullname}
    \bibliography{main}
}

\appendix
\onecolumn
\section{Qualitative Examples}

\begin{longtable}{rcccc}
\caption{The attention map visualization of the ViT-L/16 models trained on different scales of data. Images are selected to represent cultures in Western-centric countries and countries where low-resource languages are spoken.}
\label{tab:attention_maps} \\

\toprule
\textbf{Concept} & \textbf{Image} & \textbf{1B} & \textbf{10B} & \textbf{100B} \\
\midrule
\endfirsthead  

\toprule
\textbf{Concept} & \textbf{Image} & \textbf{1B} & \textbf{10B} & \textbf{100B} \\
\midrule
\endhead  

\bottomrule
\endfoot  

\bottomrule
\endlastfoot 

Street (New York) \footnote{By Terabass, CC BY-SA 3.0, https://commons.wikimedia.org/w/index.php?curid=134418052} &
\includegraphics[width=0.14\linewidth,height=0.14\linewidth]{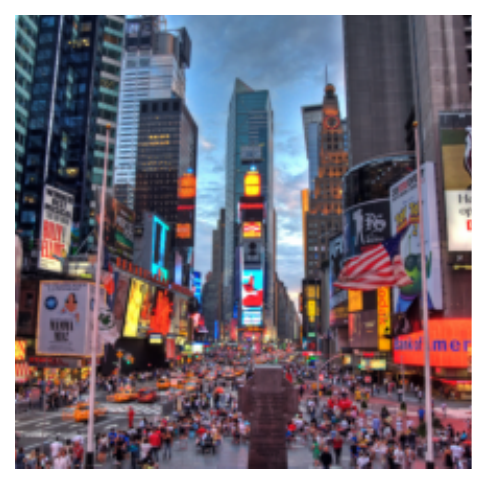} &
\includegraphics[width=0.14\linewidth]{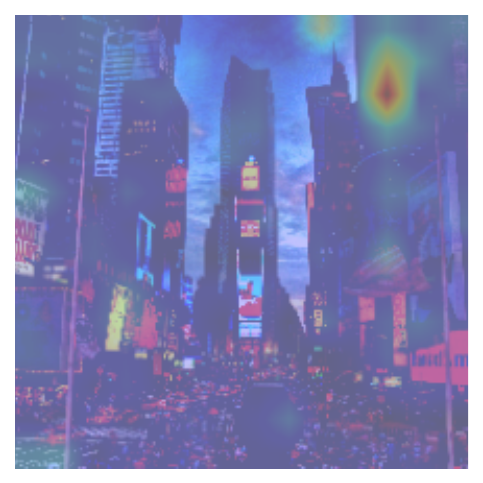} &
\includegraphics[width=0.14\linewidth]{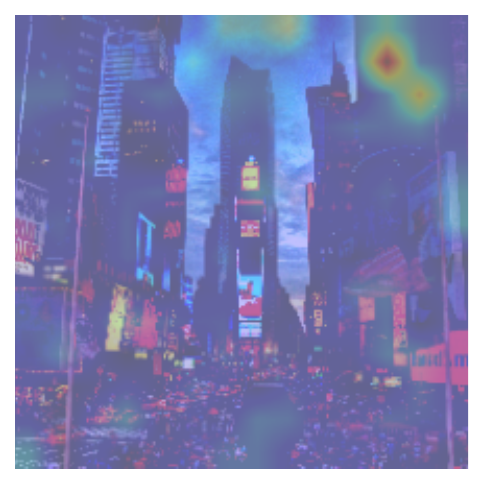} &
\includegraphics[width=0.14\linewidth]{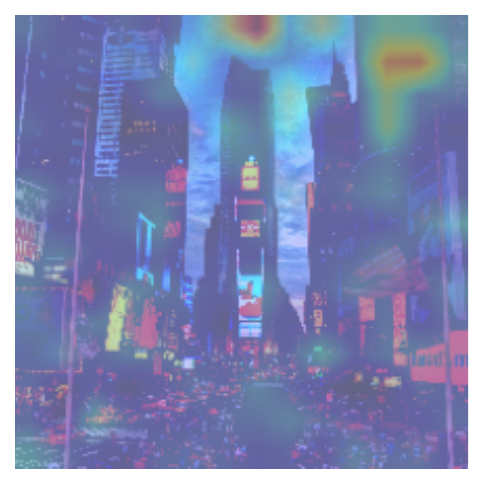} \\

\hline
Pub (London) \footnote{By Ricardalovesmonuments - Own work, CC BY-SA 4.0, https://commons.wikimedia.org/w/index.php?curid=122810839} &
\includegraphics[width=0.14\linewidth,height=0.14\linewidth]{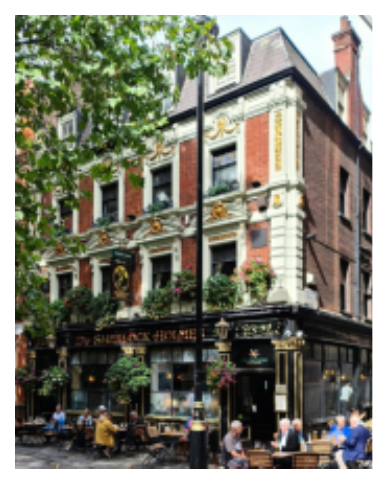} &
\includegraphics[width=0.14\linewidth]{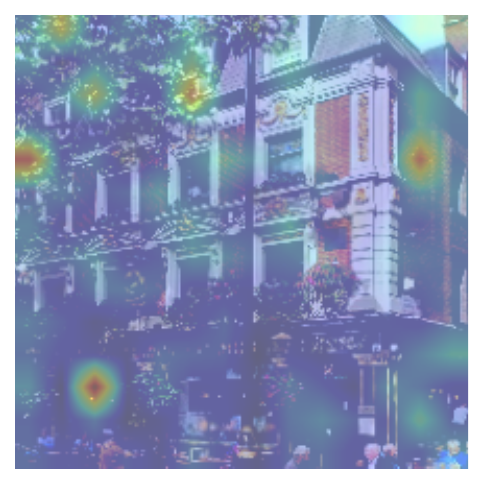} &
\includegraphics[width=0.14\linewidth]{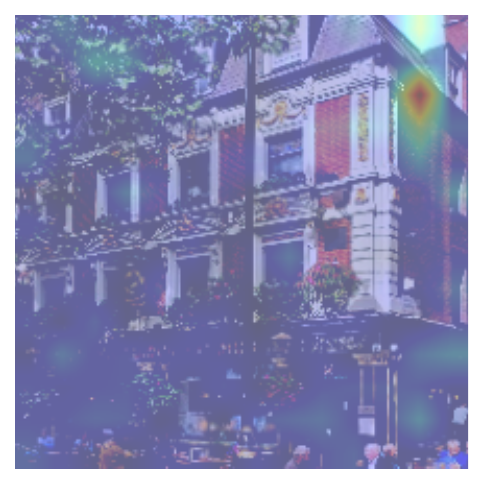} &
\includegraphics[width=0.14\linewidth]{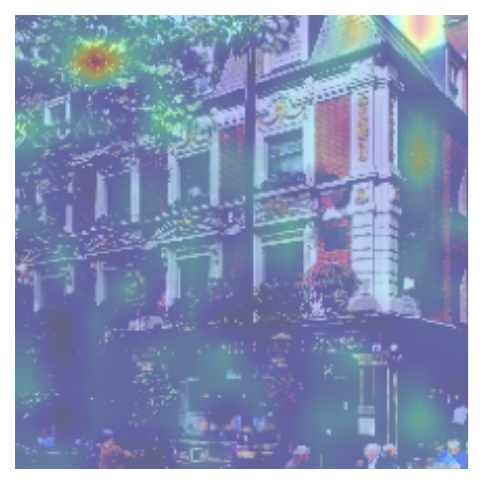} \\

\hline
Bison (Yellowstone) \footnote{Source: Yellowstone National Park, https://www.yellowstonenationalparklodges.com/connect/yellowstone-hot-spot/yellowstone-where-the-bison-roam/} &
\includegraphics[width=0.14\linewidth,height=0.14\linewidth]{figures/image_old_bison_yellowstone.pdf} &
\includegraphics[width=0.14\linewidth]{figures/attn_map_old_bison_yellowstone_SigLIP_webli1b_L16.pdf} &
\includegraphics[width=0.14\linewidth]{figures/attn_map_old_bison_yellowstone_SigLIP_webli10b_L16.pdf} &
\includegraphics[width=0.14\linewidth]{figures/attn_map_old_bison_yellowstone_SigLIP_webli100b_L16.pdf} \\

\hline
Igorot Dance (Igorot) \footnote{Source: Itogon, https://itogon.wordpress.com/2012/04/26/book-goes-to-heart-of-igorot-people/} &
\includegraphics[width=0.14\linewidth,height=0.14\linewidth]{figures/image_old_panagbenga.pdf} &
\includegraphics[width=0.14\linewidth]{figures/attn_map_old_panagbenga_SigLIP_webli1b_L16.pdf} &
\includegraphics[width=0.14\linewidth]{figures/attn_map_old_panagbenga_SigLIP_webli10b_L16.pdf} &
\includegraphics[width=0.14\linewidth]{figures/attn_map_old_panagbenga_SigLIP_webli100b_L16.pdf} \\

\hline
Kathputli Kala Chitra (Hindi) \footnote{Source: The Better India, https://thebetterindia.com/57220/journey-indian-handicraft-landscape/} &
\includegraphics[width=0.14\linewidth,height=0.14\linewidth]{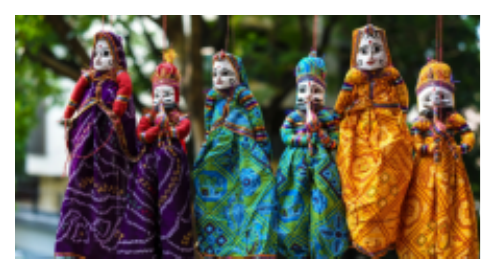} &
\includegraphics[width=0.14\linewidth]{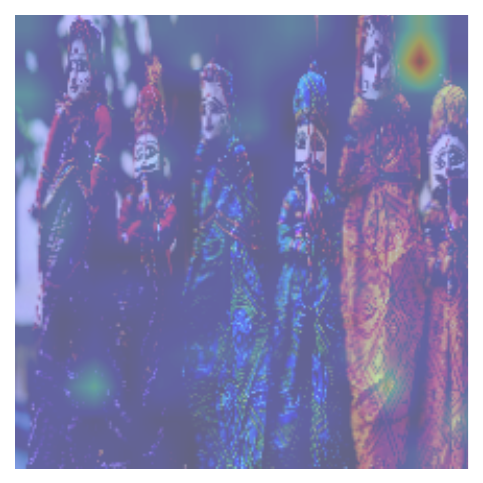} &
\includegraphics[width=0.14\linewidth]{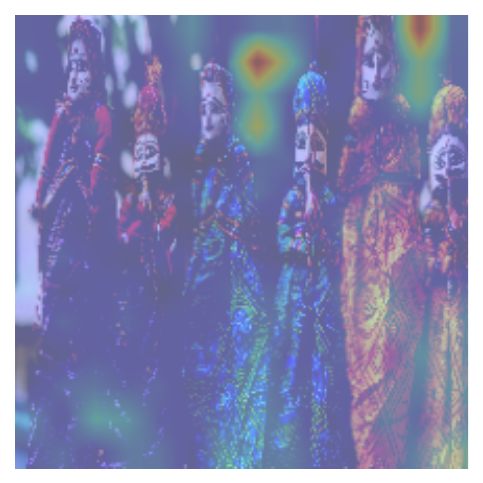} &
\includegraphics[width=0.14\linewidth]{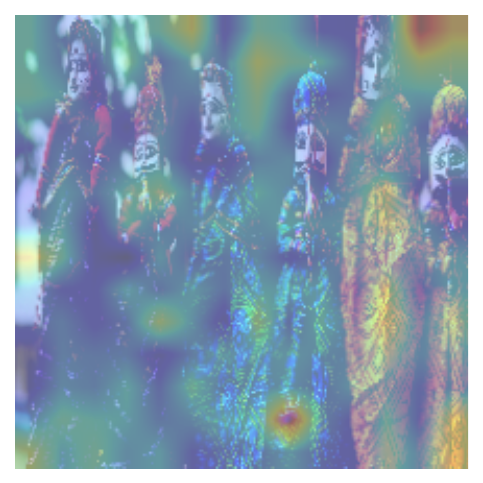} \\

\hline
Igloo (Inuit) \footnote{Drawn by unknown artist based on sketches by C.F. Hall and photographed from the book by User:Finetooth - Arctic Researches and Life Among the Esquimaux: Being the Narrative of an Expedition in Search of Sir John Franklin in the Years 1860, 1861, and 1862 by Charles Francis Hall (1865), New York: Harper and Brothers., Public Domain, https://commons.wikimedia.org/w/index.php?curid=3648025} &
\includegraphics[width=0.14\linewidth,height=0.14\linewidth]{figures/image_igloos.pdf} &
\includegraphics[width=0.14\linewidth]{figures/attn_map_old_igloos_SigLIP_webli1b_L16.pdf} &
\includegraphics[width=0.14\linewidth]{figures/attn_map_old_igloos_SigLIP_webli10b_L16.pdf} &
\includegraphics[width=0.14\linewidth]{figures/attn_map_old_igloos_SigLIP_webli100b_L16.pdf} \\

\hline
Pohela Boishakh (Bengali) \footnote{Source: EyeNews, https://www.eyenews.news/english/Today-is-Pahela-Baishakh-the-first-day-of-Bengal-1430/757} &
\includegraphics[width=0.14\linewidth,height=0.14\linewidth]{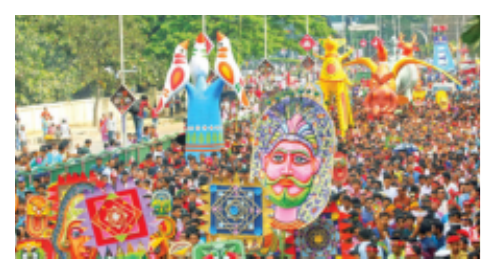} &
\includegraphics[width=0.14\linewidth]{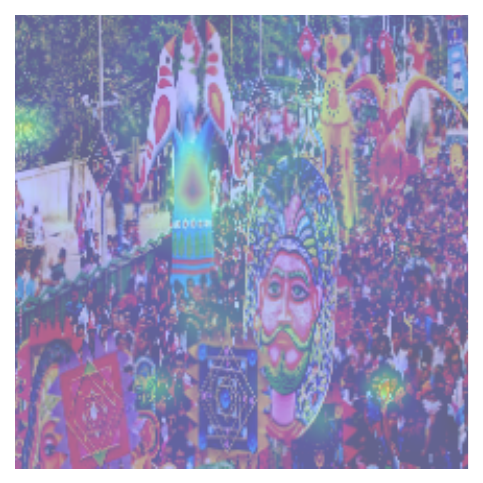} &
\includegraphics[width=0.14\linewidth]{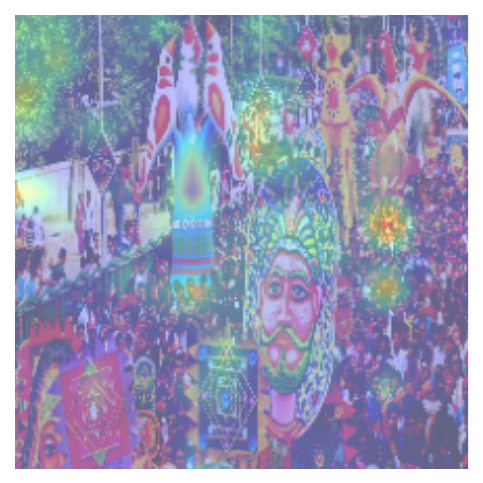} &
\includegraphics[width=0.14\linewidth]{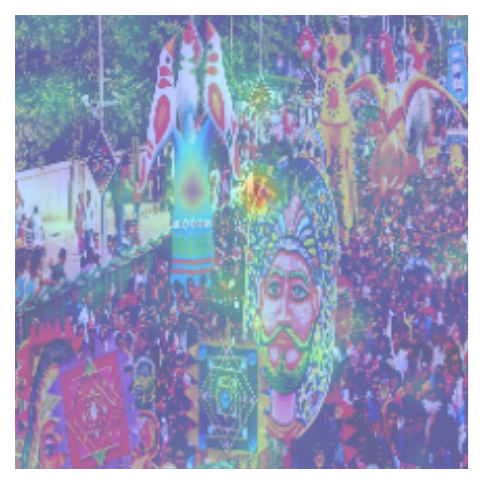} \\

\end{longtable}

\newpage

\section{Scaling Law}
\label{appendix:scaling_law}

\begin{table*}[h]
    \centering\scriptsize
    \caption{Evaluations and scaling laws on Western-centric benchmarks, where scaling from 10B to 100B examples shows limited benefits.}
    \label{tab:west_standard_setup}
    \begin{tabularx}{\columnwidth}{@{}l|l|YYY|YYY|YYY@{}}
    \toprule
    \bf Model & \bf Metric (err\%) &\multicolumn{3}{c}{\textbf{Value @ 100B ex}} &\multicolumn{6}{c}{\textbf{Scaling Laws}}\\
    && &&&\multicolumn{3}{c}{\bf exponent} &\multicolumn{3}{c}{\bf limit}\\
    &&1B &10B &100B &1B &10B &100B &1B &10B &100B\\
    \midrule
    \multicolumn{11}{c}{\em Zero-shot classification}\\[2pt]

\multirow{3}{*}{B} 
& ImageNet
& 41.2&\underline{39.4}&\bf39.0&-0.58&-0.97&-0.65&40.1&\underline{38.5}&\bf37.9
\\
& CIFAR100
&\underline{36.6}&\bf35.9&{36.8}&-0.26&-0.23&-0.24&33.8&\bf32.5&\underline{33.7}\\

& Pet 
&25.4&\underline{23.7}&\bf22.3&-0.43&-0.45&-0.37&22.3&\underline{21.7}&\bf18.4\\[2pt]
\multirow{3}{*}{L} 
& ImageNet 
&  31.2&\underline{29.7}&\bf28.5&-0.92&-0.91&-0.82&30.7&\underline{29.0}&\bf27.1
\\
& CIFAR100 
& 25.0&\underline{23.8}&\bf23.4&-0.26&-0.32&-0.43&22.7&\bf20.7&\underline{21.1}\\

& Pet 
& 14.4&\underline{12.5}&\bf9.5&-0.61&-0.57&-0.51&12.3&\underline{9.6}&\bf7.0\\[2pt]
\multirow{3}{*}{H} 
& ImageNet 
&  29.6&\underline{25.6}&\bf24.9&-0.36&-0.64&-0.52&26.7&\underline{24.5}&\bf23.3
\\
& CIFAR100 
& 23.5&\bf19.8&\underline{21.4}&-0.25&-0.36&-0.29&20.6&\underline{18.0}&\bf17.6\\
& Pet 
&10.3&\underline{7.5}&\bf7.2&-0.45&-0.42&-0.50&8.1&\underline{5.3}&\bf4.6\\\midrule
    \multicolumn{11}{c}{\em Retrieval @1}\\[2pt]
\multirow{4}{*}{B} 
& COCO I2T@1 
&  56.5&\bf51.6&\underline{53.4}&-0.24&-0.49&-0.30&52.4&\bf{49.9}&\underline{50.7}\\
& COCO T2I@1 
&  70.9&\bf68.8&\underline{70.0}&-0.34&-0.39&-0.69&69.6&\bf{67.1}&\underline{69.5}
\\
& Flickr I2T@1 
&  24.2&\underline{21.2}&\bf21.1&-0.24&-0.34&-0.23&21.5&\underline{18.1}&\bf17.0\\

& Flickr T2I@1 
&  43.1&\bf40.3&\underline{40.4}&-0.32&-0.42&-0.30&40.9&\underline{37.5}&\bf36.7\\[2pt]

\multirow{4}{*}{L} 
& COCO I2T@1 
&  49.7&\underline{47.2}&\bf45.3&-0.24&-0.41&-0.30&45.8&\underline{44.7}&\bf42.9\\

& COCO T2I@1 
&  68.2&\underline{64.3}&\bf62.5&-0.19&-0.42&-0.41&64.2&\underline{62.6}&\bf60.5\\

& Flickr I2T@1 
&  20.4&\bf15.5&\underline{16.6}&-0.21&-0.45&-0.21&16.5&\underline{14.1}&\bf13.4\\

& Flickr T2I@1 
&  39.9&\bf32.3&\underline{32.5}&-0.10&-0.42&-0.42&\underline{34.6}&\bf30.7&\bf30.7\\[2pt]

\multirow{4}{*}{H} 
& COCO I2T@1 
&  48.6&\bf42.0&\underline{42.5}&-0.21&-0.62&-0.47&44.6&\bf40.3&\underline{40.6}\\
& COCO T2I@1 
&  64.9&\underline{60.3}&\bf59.3&-0.30&-0.55&-0.43&62.8&\underline{58.9}&\bf57.3\\
& Flickr I2T@1 
&  16.8&\bf13.5&\underline{13.9}&-0.23&-0.40&-0.23&12.2&\underline{11.4}&\bf11.3\\
& Flickr T2I@1 
& 34.3&\underline{28.5}&\bf28.0&-0.23&-0.56&-0.46&29.6&\underline{26.8}&\bf25.9\\ \midrule

    \multicolumn{11}{c}{\em 10-shot}\\[2pt]
\multirow{8}{*}{B} 
& Imagenet 
& 46.6&\underline{45.6}&\bf44.7&-0.82&-0.61&-0.49&46.2&\underline{44.4}&\bf43.3\\

& Birds 
&\underline{53.8}&\bf53.5&{53.9}&-0.34&-0.40&-0.51&\bf51.5&\underline{51.6}&{52.8}\\

& Caltech 
& 8.4&\underline{8.3}&\bf8.2&-0.30&-0.24&-0.23&\underline{7.1}&7.2&\bf6.8\\

& Cars 
& 18.3&\bf16.8&\underline{17.6}&-0.63&-0.68&-0.60&17.1&\bf{15.5}&\underline{16.3}\\

& CIFAR100 
& \underline{38.7}&\bf38.6&{39.0}&-0.19&-0.22&-0.20&\underline{35.2}&\bf34.9&{35.9}\\

& Colorectal 
& \bf26.5&29.2&\underline{27.0}&-0.02&-0.06&-0.16&\bf20.2&\underline{22.6}&{24.4}\\

& Pet 
& \underline{22.9}&23.2&\bf22.1&-1.77&-0.62&-0.77&21.6&\underline{21.3}&\bf20.6\\

& DTD 
& \bf29.7&\underline{30.9}&\underline{30.9}&-0.28&-0.24&-0.19&\underline{27.9}&28.3&\bf27.2\\[3pt]

\multirow{8}{*}{L} 
& Imagenet 
&35.1&\underline{35.0}&\bf33.7&-0.67&-0.68&-0.63&34.1&\underline{34.0}&\bf32.5\\

& Birds 
& \bf44.0&45.3&\underline{44.3}&-0.51&-0.43&-0.51&\bf42.1&43.2&\underline{42.7}\\

& Caltech 
& \bf6.4&\underline{7.4}&7.5&-0.43&-0.17&-0.18&\underline{5.9}&\bf4.8&\bf4.8\\

& Cars 
& \bf11.1&\underline{11.3}&{11.5}&-0.54&-0.49&-0.41&10.1&\bf9.7&\underline{9.9}\\

& CIFAR100 
& 27.5&\underline{26.7}&\bf25.5&-0.24&-0.29&-0.41&24.0&\underline{23.7}&\bf22.9\\

& Colorectal 
&  24.0&\underline{23.5}&\bf22.6&-0.18&-0.20&-0.27&\bf18.8&\underline{20.2}&{20.5}\\

& Pet 
& \underline{12.3}&12.5&\bf11.8&-0.70&-0.65&-0.53&\underline{11.3}&11.4&\bf10.3\\

& DTD 
& 28.5&\bf27.1&\underline{27.9}&-0.22&-0.25&-0.23&\underline{25.2}&\bf25.1&{25.5}\\[3pt]

\multirow{8}{*}{H} 
& Imagenet 
&32.4&\underline{29.8}&\bf29.3&-0.41&-0.73&-0.79&30.3&\underline{29.0}&\bf28.3\\

& Birds 
&41.6&\underline{39.1}&\bf36.3&-0.67&-0.52&-0.47&40.6&\underline{37.4}&\bf33.9\\

& Caltech 
& \bf5.7&\underline{6.0}&{8.9}&-0.21&-0.08&-0.11&\underline{4.3}&\bf3.7&{4.6}\\

& Cars 
& 11.3&\underline{10.3}&\bf9.6&-0.27&-0.88&-0.44&\underline{9.1}&10.1&\bf8.3\\

& CIFAR100 
& 25.8&\bf23.8&\underline{24.2}&-0.22&-0.25&-0.24&21.4&\underline{21.1}&\bf{19.7}\\

& Colorectal 
& \bf25.2&26.2&\underline{25.9}&-0.22&-0.20&-0.15&\underline{19.7}&\bf17.9&{20.7}\\

& Pet 
& 10.8&\underline{9.1}&\bf8.7&-0.92&-0.48&-0.46&10.3&\underline{7.6}&\bf6.5\\

& DTD 
& 29.2&\bf26.1&\underline{26.8}&-0.16&-0.23&-0.23&25.0&\bf23.8&\underline{24.8}\\
 \bottomrule
    \end{tabularx}
\end{table*}
\begin{table*}[h]
    \centering\scriptsize
    \caption{Evaluations and scaling laws on culture diversity benchmarks, where scaling from 10B to 100B examples shows larger benefits.}
    \label{tab:culture_standard_setup}
    \begin{tabularx}{\columnwidth}{@{}l|l|YYY|YYY|YYY@{}}
    \toprule
    \bf Model & \bf Metric (err \%) &\multicolumn{3}{c}{\textbf{Value @ 100B ex}} &\multicolumn{6}{c}{\textbf{Scaling Laws}}\\
    && &&&\multicolumn{3}{c}{\bf exponent} &\multicolumn{3}{c}{\bf limit}\\
    &&1B &10B &100B &1B &10B &100B &1B &10B &100B\\
    \midrule
    \multicolumn{11}{c}{\em 10-shot Geolocalization}\\[2pt]

\multirow{3}{*}{B} 
& Dollar Street
& 77.7&\underline{75.8}&\bf72.1&-0.38&-0.36&-0.37&76.3&\underline{73.7}&\bf70.2\\
& GeoDE-Country
&72.8&\underline{71.5}&\bf71.4&-0.35&-0.31&-0.37&70.8&\underline{69.6}&\bf68.9\\
&GeoDE-Region & 61.1&\underline{60.8}&\bf59.2&-0.26&-0.22&-0.29&58.8&\bf57.0&\underline{57.3}\\[2pt]

\multirow{3}{*}{L} 
& Dollar Street 
& \underline{63.6}&64.1&\bf58.3&-1.09&-0.38&-0.94&63.2&\underline{60.1}&\bf57.5\\

& GeoDE-Country 
& \underline{61.9}&62.3&\bf57.8&-0.40&-0.30&-1.11&58.8&\underline{58.0}&\bf56.6\\

& GeoDE-Region  
& 54.2&\underline{53.6}&\bf48.3&-0.15&-0.16&-0.39&49.9&\underline{46.9}&\bf46.3\\[2pt]

\multirow{3}{*}{H} 
& Dollar Street 
& 64.6&\underline{59.1}&\bf53.7&-0.30&-0.56&-0.64&61.0&\underline{56.4}&\bf52.5\\
& GeoDE-Country 
&56.9&\underline{50.2}&\bf47.6&-0.23&-0.78&-0.62&52.2&\underline{49.4}&\bf46.1\\
& GeoDE-Region  
&54.6&\underline{47.6}&\bf44.7&0.00&-0.38&-0.31&50.1&\underline{45.3}&\bf41.0\\\midrule

    \multicolumn{11}{c}{\em Zero-shot classification}\\[2pt]
\multirow{4}{*}{B} 
& Dollar Street 
& 52.0&\underline{51.9}&\bf51.6&-0.38&-0.25&-0.28&\underline{50.4}&\bf49.7&\bf49.7\\

& GeoDE
&\bf7.8&\underline{8.3}&{8.7}&-0.24&-0.26&-0.25&\underline{6.1}&6.7&\bf5.4\\

& GLDv2
& 65.0&\underline{61.0}&\bf59.4&-0.46&-0.72&-0.51&61.6&\underline{59.3}&\bf56.8\\[2pt]

\multirow{4}{*}{L} 
& Dollar Street
&50.2&\bf48.1&\underline{49.0}&-0.22&-0.35&-0.17&\underline{46.9}&\bf46.2&\bf46.2\\

& GeoDE 
& 6.0&\underline{5.9}&\bf4.9&-0.29&-0.17&-0.25&4.7&\underline{4.3}&\bf3.3\\

& GLDv2 
& 50.4&\underline{46.4}&\bf45.7&-0.53&-0.93&-0.89&48.5&\underline{44.8}&\bf44.1\\[2pt]

\multirow{4}{*}{H} 
& Dollar Street
&50.0&\underline{48.6}&\bf47.4&-0.15&-0.13&-0.20&43.9&\underline{44.2}&\bf44.1\\
& GeoDE
&  6.0&\underline{4.9}&\bf4.8&-0.19&-0.22&-0.24&\bf3.3&\bf3.3&\underline{3.5}\\
& GLDv2 
&48.1&\underline{40.1}&\bf38.8&-0.52&-1.34&-0.80&46.0&\underline{39.0}&\bf36.8\\ \bottomrule
    \end{tabularx}
\end{table*}

\FloatBarrier
\newpage

\section{Association Bias}
\label{appendix:ab}

\begin{figure*}[h!]
    \centering
    \includegraphics[width=0.32\columnwidth]{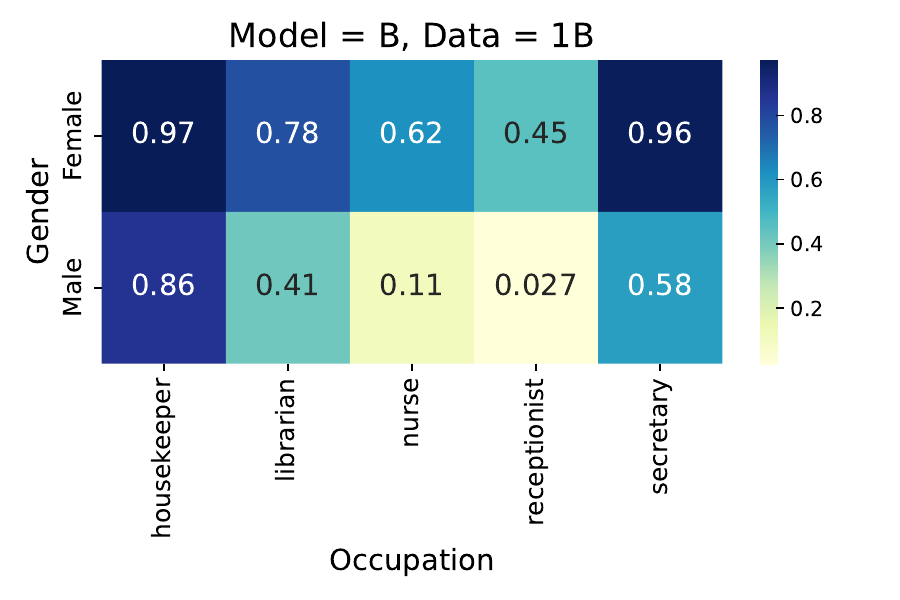}
    \includegraphics[width=0.32\columnwidth]{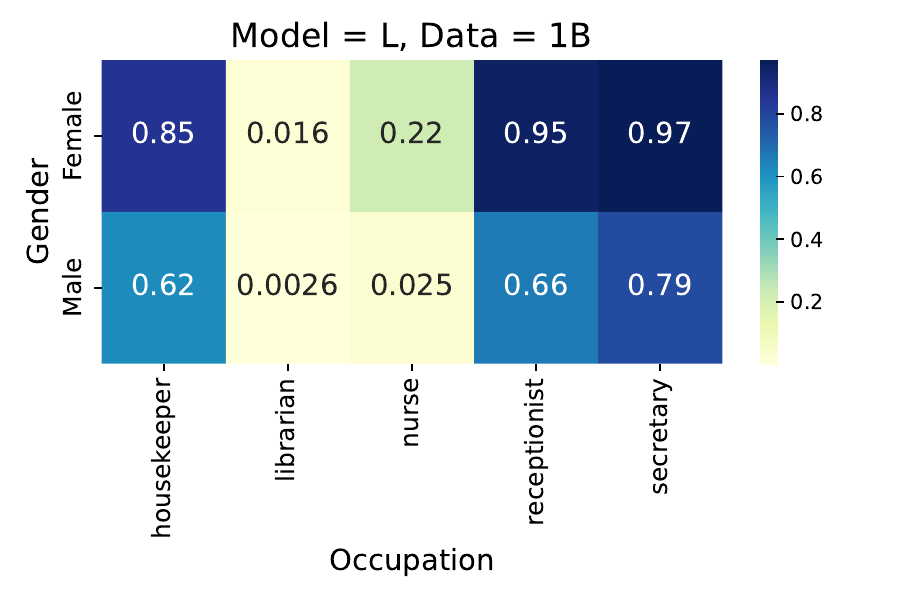}
    \includegraphics[width=0.32\columnwidth]{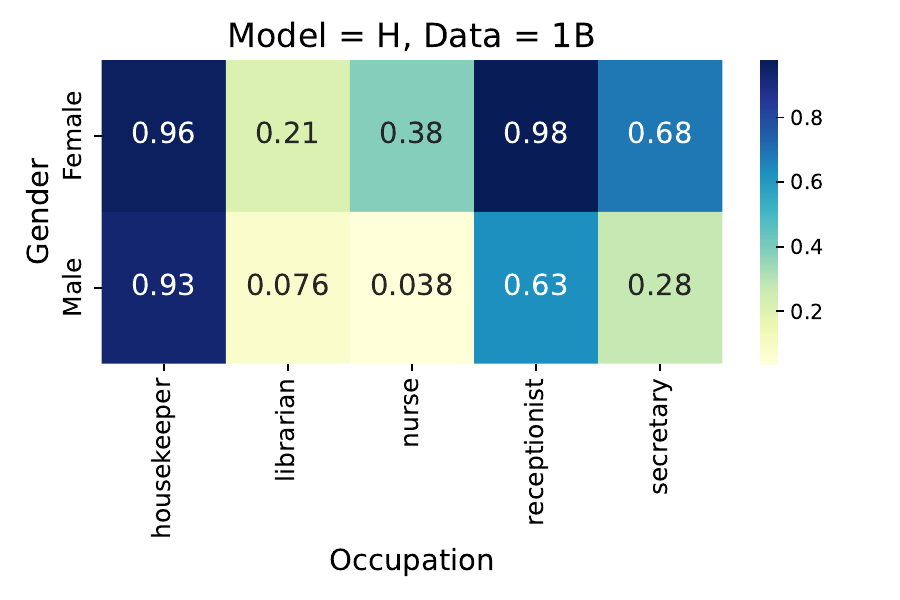}\\
    \includegraphics[width=0.32\columnwidth]{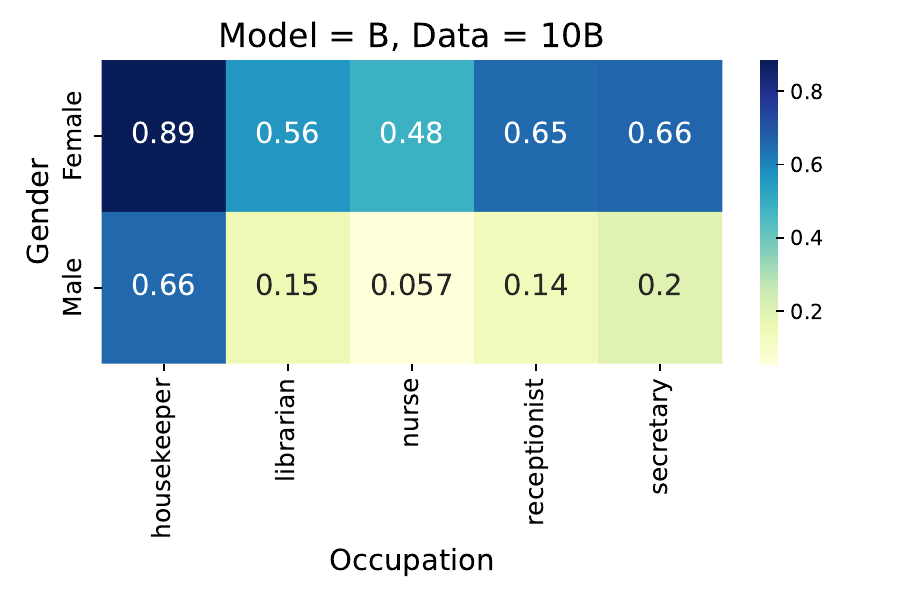}
    \includegraphics[width=0.32\columnwidth]{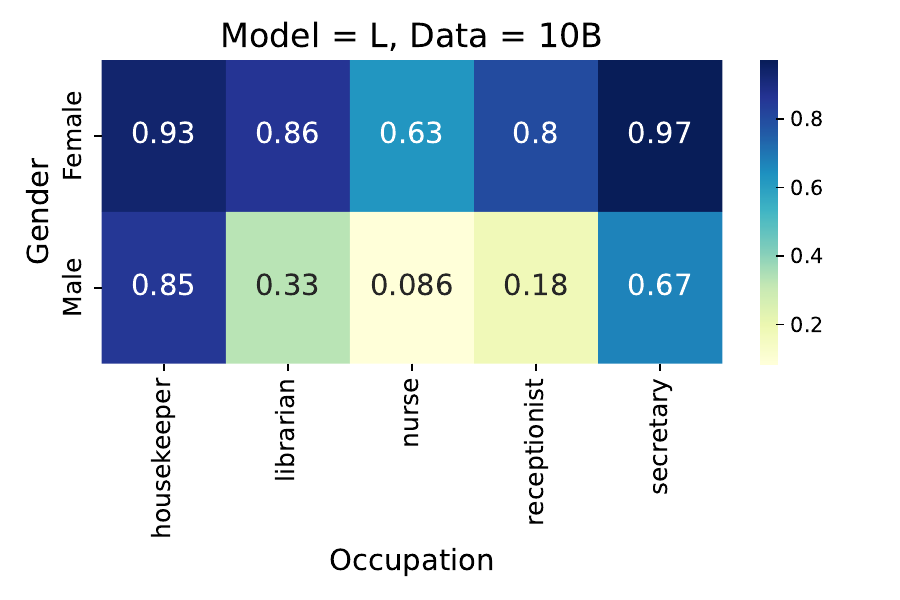}
    \includegraphics[width=0.32\columnwidth]{figures/fairness/ab_heatmap_H_10B.pdf}\\
    \includegraphics[width=0.32\columnwidth]{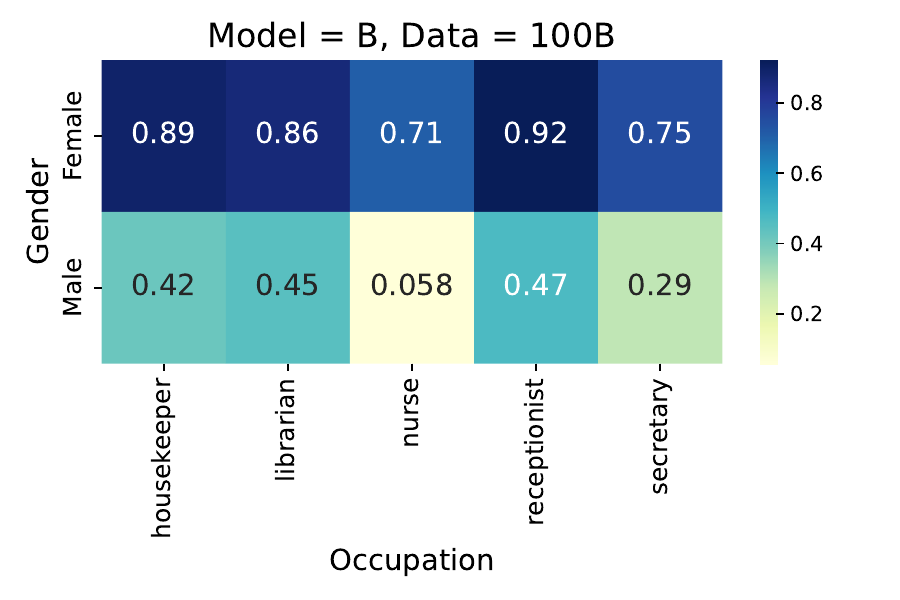}
    \includegraphics[width=0.32\columnwidth]{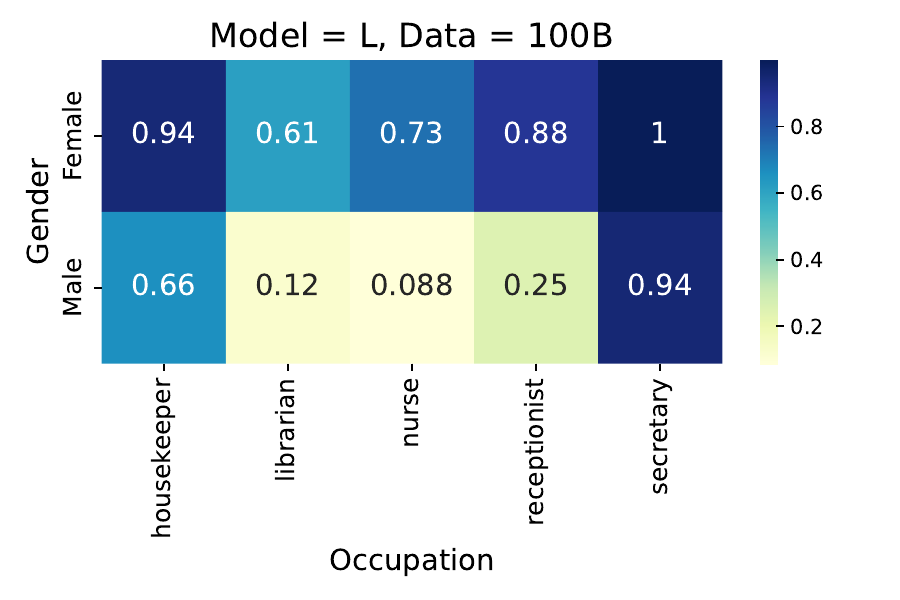}
    \includegraphics[width=0.32\columnwidth]{figures/fairness/ab_heatmap_H_100B.pdf}
    \caption{Association bias between gender and occupation, evaluated in scaled models and data.}
    \label{fig:ab_app}
\end{figure*}

\section{Performance Disparity}\label{appendix:perf_disp}
\begin{table*}[h]
    \centering\scriptsize
    \caption{Performance disparity results for various SigLIP models pretrained on 100 billion seen examples of 1B, 10B, and 100B datasets. Here, disparity corresponds to the maximum gap across subgroups in Dollar Street (by income level) and GeoDE (by geographic region). Pretraining on 100B examples tends to improve disparity overall.}
    \label{tab:perf_disparity}
    \begin{tabularx}{\columnwidth}{ll|YYYYYY|Y}
    \toprule
    \bf Model & \bf Data Scale &\multicolumn{6}{c}{\bf Performance per Subgroup} & \bf Disparity\\ \midrule
    \multicolumn{8}{c}{\em 0-shot Dollar Street}\\[2pt]
    & & \bf 0-200	& \bf 200-685	& \bf 685-1998	& \bf $>$1998
    & & & \\ \midrule
B&1B&29.4&43.9&56.5&62.0&&&32.5\\
B&10B&31.6&44.0&55.4&61.5&&&29.9\\
B&100B&32.0&44.3&56.3&61.0&&&\bf29.0\\[3pt]
L&1B&33.7&44.7&57.3&63.4&&&\bf29.7\\
L&10B&35.7&47.8&58.7&65.5&&&29.8\\
L&100B&33.7&46.6&59.5&64.1&&&30.4\\[3pt]
H&1B&32.3&44.9&58.4&64.5&&&32.2\\
H&10B&33.9&46.3&58.6&66.9&&&33.0\\
H&100B&34.1&48.2&62.2&66.1&&&\bf32.1\\ \midrule

    \multicolumn{8}{c}{\em 0-shot GeoDE}\\[2pt]
    & & \bf Africa	& \bf Americas	& \bf East-Asia	& \bf Europe & \bf South-East Asia & \bf West Asia
    & \\ \midrule
B&1B&89.4&92.1&91.8&94.1&92.5&93.4&4.7\\
B&10B&88.4&91.8&91.4&94.0&92.2&93.0&5.5\\
B&100B&88.8&91.4&91.0&93.3&91.7&92.2&\bf4.4\\[3pt]
L&1B&92.0&94.0&94.0&95.2&94.2&94.9&3.2\\
L&10B&91.8&94.4&94.0&95.8&94.2&94.7&4.0\\
L&100B&93.5&95.1&95.4&96.2&95.0&95.8&\bf2.8\\[3pt]
H&1B&91.5&94.4&94.7&95.2&94.1&94.5&3.6\\
H&10B&93.4&95.4&95.0&96.5&95.1&95.6&3.0\\
H&100B&93.6&95.1&95.3&96.3&95.2&95.8&\bf2.7\\

 \bottomrule
    \end{tabularx}
\end{table*}
\newpage

\section{Evaluations of Data Scaling}
\label{appendix:data_scale}

{\scriptsize

\newcommand{\dottedline}{\multicolumn{11}{c}{\dotfill} \\}
\scriptsize
\begin{longtable}{l|l|rrr|rrr|rrr} 
\caption{Detailed evaluation results of ViT-B/L/H models on 1/10/100 billion scale datasets. All metrics are measured by error rate, with the exception of ``Representation Bias'', which is measured by disparity.}
\label{tab:data_scale} \\

\toprule
& & \multicolumn{3}{c}{ViT-B/16} & \multicolumn{3}{|c|}{ViT-L/16} & \multicolumn{3}{c}{ViT-H/14} \\
\cmidrule(lr){3-5}\cmidrule(lr){6-8}\cmidrule(lr){9-11}
Metric & Category & 1B & 10B & 100B & 1B & 10B & 100B & 1B & 10B & 100B \\ 
\midrule
\endfirsthead  

\endhead  

\endfoot  

\bottomrule
\endlastfoot 

ImageNet 0-shot Classification & \multirow{15}{*}{Western} & 41.21 & 39.35 & 39.04 & 31.23 & 29.70 & 28.49 & 29.60 & 25.60 & 24.90 \\
Cifar100 0-shot Classification &  & 36.62 & 35.87 & 36.80 & 25.02 & 23.75 & 23.36 & 23.49 & 19.79 & 21.42 \\
Pet 0-shot Classification & & 25.40 & 23.71 & 22.27 & 14.36 & 12.46 & 9.46 & 10.33 & 7.47 & 7.17 \\
ImageNet 10-shot Classification & & 46.65 & 45.63 & 44.74 & 35.11 & 34.95 & 33.71 & 32.44 & 29.76 & 29.34 \\
Cifar100 10-shot Classification & & 38.73 & 38.63 & 39.02 & 27.50 & 26.70 & 25.49 & 25.76 & 23.79 & 24.21 \\
Pet 10-shot Classification & & 22.95 & 23.19 & 22.08 & 12.32 & 12.48 & 11.80 & 10.85 & 9.13 & 8.67 \\
Bird 10-shot Classification & & 53.80 & 53.47 & 53.90 & 44.05 & 45.25 & 44.29 & 41.65 & 39.13 & 36.31 \\
Caltech 10-shot Classification & & 8.37 & 8.33 & 8.23 & 6.41 & 7.40 & 7.53 & 5.70 & 6.02 & 8.93 \\
Cars 10-shot Classification & & 18.29 & 16.79 & 17.60 & 11.14 & 11.33 & 11.47 & 11.32 & 10.30 & 9.60 \\
Colorectal Histology 10-shot Classification & & 26.53 & 29.23 & 27.00 & 24.00 & 23.53 & 22.57 & 25.17 & 26.17 & 25.87 \\
DTD 10-shot Classification & & 29.73 & 30.85 & 30.90 & 28.46 & 27.07 & 27.93 & 29.20 & 26.12 & 26.76 \\
COCO Image-Text 0-shot Retrieval & & 56.46 & 51.62 & 53.44 & 49.70 & 47.18 & 45.28 & 48.62 & 42.04 & 42.48 \\
COCO Text-Image 0-shot Retrieval & & 70.90 & 68.84 & 70.01 & 68.16 & 64.32 & 62.51 & 64.86 & 60.32 & 59.29 \\
Flickr Image-Text 0-shot Retrieval & & 24.20 & 21.20 & 21.10 & 20.40 & 15.50 & 16.60 & 16.80 & 13.50 & 13.90 \\
Flickr Text-Image 0-shot Retrieval & & 43.12 & 40.26 & 40.42 & 39.94 & 32.32 & 32.52 & 34.26 & 28.46 & 28.00 \\
\dottedline
Dollar Street 0-shot Classification & \multirow{6}{*}{Culture} & 52.04 & 51.88 & 51.60 & 50.23 & 48.10 & 49.03 & 50.00 & 48.58 & 47.35 \\
Dollar Street 10-shot Classification & & 77.69 & 75.81 & 72.12 & 63.56 & 64.09 & 58.29 & 64.60 & 59.10 & 53.69 \\
GeoDE 0-shot Classification & & 7.85 & 8.27 & 8.65 & 6.01 & 5.90 & 4.88 & 5.99 & 4.87 & 4.81 \\
GeoDE (country) 10-shot Classification & & 72.75 & 71.47 & 71.36 & 61.94 & 62.31 & 57.85 & 56.94 & 50.22 & 47.55 \\
GeoDE (region) 10-shot Classification & & 61.09 & 60.80 & 59.18 & 54.21 & 53.59 & 48.29 & 54.56 & 47.63 & 44.68 \\
GLDv2 0-shot Classification & & 65.05 & 60.96 & 59.40 & 50.39 & 46.37 & 45.72 & 48.05 & 40.08 & 38.78 \\
\dottedline
Representation Bias & \multirow{13}{*}{Fairness} & 33.15 & 34.54 & 35.21 & 38.18 & 36.35 & 35.51 & 36.76 & 35.01 & 36.61 \\
Income 0-200 Classification & & 70.57 & 68.43 & 67.97 & 66.30 & 64.35 & 66.30 & 67.69 & 66.11 & 65.92 \\
Income 200-285 Classification & & 56.07 & 55.98 & 55.70 & 55.33 & 52.18 & 53.38 & 55.14 & 53.66 & 51.81 \\
Income 285-685 Classification & & 43.45 & 44.57 & 43.73 & 42.71 & 41.32 & 40.48 & 41.60 & 41.41 & 37.79 \\
Income \textgreater1998 Classification & & 38.05 & 38.51 & 38.98 & 36.56 & 34.51 & 35.91 & 35.53 & 33.12 & 33.86 \\
GeoDE: Africa & & 10.58 & 11.56 & 11.15 & 7.99 & 8.24 & 6.55 & 8.46 & 6.56 & 6.40 \\
GeoDE: Americas & & 7.94 & 8.16 & 8.58 & 6.03 & 5.57 & 4.92 & 5.60 & 4.57 & 4.86 \\
GeoDE: EastAsia & & 8.15 & 8.57 & 8.99 & 5.98 & 5.96 & 4.56 & 5.30 & 5.01 & 4.68 \\
GeoDE: Europe & & 5.92 & 6.02 & 6.75 & 4.81 & 4.20 & 3.75 & 4.83 & 3.53 & 3.75 \\
GeoDE: SouthEastAsia & & 7.51 & 7.81 & 8.26 & 5.78 & 5.78 & 5.02 & 5.86 & 4.89 & 4.76 \\
GeoDE: WestAsia & & 6.57 & 7.01 & 7.85 & 5.11 & 5.30 & 4.19 & 5.50 & 4.42 & 4.19 \\
Perceived Gender & & 9.05 & 8.34 & 8.11 & 5.25 & 6.06 & 4.97 & 5.53 & 4.59 & 4.60 \\
Perceived Race & & 41.44 & 40.52 & 46.87 & 44.57 & 46.88 & 46.04 & 45.47 & 45.67 & 49.84 \\
\dottedline
XM3600 Image-Text: Arabic & \multirow{22}{*}{Multiling} & 61.78 & 53.42 & 53.36 & 53.58 & 45.00 & 44.56 & 52.25 & 41.64 & 41.00 \\
XM3600 Image-Text: Bengali & & 95.69 & 80.64 & 77.06 & 90.81 & 66.36 & 63.75 & 88.17 & 61.22 & 56.69 \\
XM3600 Image-Text: Czech & & 60.78 & 51.89 & 50.83 & 52.31 & 43.81 & 42.22 & 49.94 & 40.11 & 39.44 \\
XM3600 Image-Text: Danish & & 55.58 & 45.39 & 45.75 & 45.08 & 35.06 & 31.00 & 43.03 & 29.92 & 28.75 \\
XM3600 Image-Text: German & & 39.47 & 31.53 & 31.78 & 30.61 & 24.28 & 24.03 & 29.17 & 22.75 & 21.89 \\
XM3600 Image-Text: Greek & & 74.36 & 63.00 & 61.86 & 67.86 & 53.64 & 50.14 & 65.67 & 49.50 & 47.33 \\
XM3600 Image-Text: English & & 56.53 & 55.03 & 55.50 & 54.14 & 52.42 & 51.67 & 53.22 & 51.42 & 49.64 \\
XM3600 Image-Text: Spanish & & 49.17 & 42.94 & 44.22 & 41.56 & 38.44 & 35.81 & 40.03 & 33.89 & 34.28 \\
XM3600 Image-Text: Persian & & 58.94 & 51.17 & 51.58 & 49.64 & 38.97 & 40.17 & 46.61 & 33.72 & 34.06 \\
XM3600 Image-Text: Finnish & & 70.64 & 53.83 & 53.61 & 59.25 & 42.67 & 39.06 & 57.39 & 34.83 & 32.86 \\
XM3600 Image-Text: Filipino & & 87.86 & 82.06 & 81.92 & 82.72 & 72.86 & 71.36 & 81.31 & 66.14 & 63.03 \\
XM3600 Image-Text: French & & 47.08 & 38.92 & 39.06 & 39.08 & 31.78 & 29.92 & 36.58 & 28.53 & 28.19 \\
XM3600 Image-Text: Hindi & & 83.53 & 74.78 & 72.39 & 77.67 & 65.67 & 63.47 & 76.92 & 62.33 & 60.64 \\
XM3600 Image-Text: Croatian & & 64.53 & 53.28 & 51.33 & 53.08 & 37.94 & 35.78 & 47.81 & 32.44 & 30.44 \\
XM3600 Image-Text: Hungarian & & 64.50 & 49.06 & 47.53 & 53.81 & 38.64 & 34.42 & 51.22 & 32.67 & 30.36 \\
XM3600 Image-Text: Indonesian & & 44.81 & 38.14 & 37.08 & 35.83 & 28.47 & 28.53 & 33.39 & 24.86 & 25.33 \\
XM3600 Image-Text: Italian & & 48.58 & 41.00 & 40.86 & 38.42 & 33.33 & 30.97 & 36.47 & 29.64 & 28.89 \\
XM3600 Image-Text: Hebrew & & 67.06 & 50.28 & 49.86 & 56.75 & 39.44 & 35.72 & 52.03 & 33.86 & 30.81 \\
XM3600 Image-Text: Japanese & & 67.36 & 55.67 & 55.42 & 59.00 & 45.42 & 44.97 & 58.47 & 42.22 & 37.94 \\
XM3600 Image-Text: Korean & & 58.64 & 49.61 & 49.53 & 50.75 & 40.33 & 38.31 & 46.81 & 35.39 & 35.08 \\
XM3600 Image-Text: Maori & & 99.61 & 99.50 & 99.42 & 99.58 & 99.22 & 99.25 & 99.31 & 98.92 & 99.17 \\
XM3600 Image-Text: Dutch & & 53.97 & 47.47 & 48.78 & 47.11 & 41.14 & 38.39 & 44.56 & 38.06 & 37.44 \\
XM3600 Image-Text: Norwegian & \multirow{50}{*}{Multiling} & 56.56 & 46.78 & 47.89 & 45.33 & 36.11 & 34.28 & 43.39 & 31.81 & 30.19 \\
XM3600 Image-Text: Polish & & 53.97 & 44.89 & 44.22 & 45.97 & 35.50 & 34.11 & 41.75 & 33.00 & 31.06 \\
XM3600 Image-Text: Portuguese & & 51.03 & 44.19 & 44.39 & 43.33 & 36.03 & 34.56 & 41.14 & 32.69 & 32.28 \\
XM3600 Image-Text: Quechua & & 95.53 & 94.08 & 93.89 & 94.64 & 93.53 & 93.92 & 94.58 & 93.06 & 92.78 \\
XM3600 Image-Text: Romanian & & 64.56 & 51.39 & 52.03 & 52.19 & 38.31 & 35.39 & 47.92 & 32.36 & 30.11 \\
XM3600 Image-Text: Russian & & 51.56 & 42.36 & 42.28 & 42.78 & 35.14 & 33.22 & 41.19 & 31.97 & 30.31 \\
XM3600 Image-Text: Swedish & & 54.03 & 44.25 & 45.69 & 44.50 & 34.94 & 34.78 & 40.69 & 31.14 & 30.78 \\
XM3600 Image-Text: Swahili & & 92.14 & 88.17 & 88.72 & 89.94 & 81.33 & 79.47 & 88.92 & 76.86 & 74.14 \\
XM3600 Image-Text: Telugu & & 98.06 & 87.08 & 80.53 & 96.08 & 76.67 & 69.69 & 96.36 & 73.08 & 65.31 \\
XM3600 Image-Text: Thai & & 79.33 & 68.67 & 67.47 & 72.61 & 59.47 & 58.86 & 71.25 & 56.86 & 52.78 \\
XM3600 Image-Text: Turkish & & 60.33 & 50.03 & 50.06 & 52.78 & 40.72 & 39.72 & 48.56 & 36.56 & 34.94 \\
XM3600 Image-Text: Ukrainian & & 62.39 & 52.25 & 49.78 & 55.19 & 41.25 & 37.83 & 52.75 & 36.94 & 33.25 \\
XM3600 Image-Text: Vietnamese & & 54.31 & 45.33 & 45.22 & 43.19 & 34.00 & 32.44 & 40.75 & 29.06 & 29.08 \\
XM3600 Image-Text: Chinese & & 63.92 & 51.08 & 51.19 & 53.67 & 42.47 & 42.50 & 54.17 & 40.53 & 38.42 \\
XM3600 Text-Image: Arabic & & 73.77 & 67.79 & 68.49 & 67.49 & 59.74 & 59.86 & 65.87 & 56.22 & 54.91 \\
XM3600 Text-Image: Bengali & & 97.19 & 89.25 & 89.53 & 95.17 & 79.72 & 77.31 & 94.22 & 76.36 & 72.42 \\
XM3600 Text-Image: Czech & & 71.81 & 64.49 & 65.48 & 65.52 & 58.57 & 58.18 & 63.59 & 55.79 & 55.07 \\
XM3600 Text-Image: Danish & & 68.23 & 59.97 & 61.73 & 60.01 & 51.18 & 49.50 & 56.72 & 46.53 & 45.46 \\
XM3600 Text-Image: German & & 55.15 & 47.80 & 49.18 & 45.85 & 39.88 & 39.75 & 43.80 & 36.56 & 36.99 \\
XM3600 Text-Image: Greek & & 82.61 & 75.69 & 75.71 & 77.96 & 69.11 & 67.35 & 75.68 & 65.45 & 64.10 \\
XM3600 Text-Image: English & & 62.32 & 59.41 & 60.78 & 58.97 & 57.57 & 56.32 & 58.15 & 56.40 & 55.82 \\
XM3600 Text-Image: Spanish & & 57.35 & 52.74 & 55.49 & 52.64 & 49.06 & 48.31 & 51.24 & 47.27 & 46.62 \\
XM3600 Text-Image: Persian & & 71.80 & 65.18 & 65.58 & 62.65 & 55.06 & 56.09 & 59.79 & 52.93 & 49.69 \\
XM3600 Text-Image: Finnish & & 81.00 & 70.80 & 68.28 & 72.96 & 59.11 & 56.24 & 70.79 & 51.07 & 49.35 \\
XM3600 Text-Image: Filipino & & 93.60 & 90.28 & 91.07 & 90.89 & 83.98 & 83.70 & 89.55 & 80.61 & 77.92 \\
XM3600 Text-Image: French & & 56.70 & 50.23 & 50.57 & 48.33 & 43.31 & 42.10 & 47.52 & 40.48 & 39.96 \\
XM3600 Text-Image: Hindi & & 91.01 & 86.55 & 86.09 & 87.43 & 81.38 & 80.01 & 87.71 & 79.21 & 78.22 \\
XM3600 Text-Image: Croatian & & 75.52 & 67.53 & 66.85 & 66.68 & 54.42 & 54.22 & 63.21 & 50.71 & 48.53 \\
XM3600 Text-Image: Hungarian & & 74.24 & 63.83 & 63.53 & 66.49 & 53.73 & 50.75 & 64.26 & 48.31 & 45.72 \\
XM3600 Text-Image: Indonesian & & 60.08 & 52.90 & 53.96 & 50.28 & 44.05 & 43.97 & 49.27 & 41.45 & 40.81 \\
XM3600 Text-Image: Italian & & 57.90 & 51.51 & 52.08 & 47.96 & 42.80 & 42.60 & 48.03 & 40.62 & 40.34 \\
XM3600 Text-Image: Hebrew & & 76.50 & 64.76 & 62.76 & 69.11 & 56.25 & 54.14 & 65.88 & 51.49 & 49.99 \\
XM3600 Text-Image: Japanese & & 76.74 & 69.20 & 68.99 & 69.56 & 62.34 & 58.44 & 69.16 & 57.06 & 54.78 \\
XM3600 Text-Image: Korean & & 70.82 & 64.88 & 67.23 & 64.52 & 56.76 & 56.51 & 61.52 & 53.57 & 52.67 \\
XM3600 Text-Image: Maori & & 99.78 & 99.78 & 99.78 & 99.73 & 99.56 & 99.62 & 99.75 & 99.67 & 99.51 \\
XM3600 Text-Image: Dutch & & 63.50 & 59.25 & 59.05 & 57.41 & 52.02 & 51.48 & 55.49 & 49.88 & 49.10 \\
XM3600 Text-Image: Norwegian & & 70.36 & 63.58 & 63.44 & 61.54 & 53.81 & 52.99 & 60.04 & 49.16 & 48.20 \\
XM3600 Text-Image: Polish & & 63.73 & 57.39 & 57.71 & 56.06 & 47.92 & 47.09 & 53.28 & 45.05 & 44.49 \\
XM3600 Text-Image: Portuguese & & 62.16 & 57.16 & 57.93 & 54.54 & 49.48 & 48.72 & 52.44 & 47.48 & 46.64 \\
XM3600 Text-Image: Quechua & & 98.46 & 97.94 & 97.85 & 97.88 & 98.14 & 98.04 & 98.18 & 98.28 & 98.26 \\
XM3600 Text-Image: Romanian & & 74.48 & 65.48 & 65.11 & 65.20 & 54.05 & 52.41 & 61.69 & 48.77 & 47.09 \\
XM3600 Text-Image: Russian & & 61.65 & 53.83 & 54.17 & 53.47 & 47.58 & 45.36 & 51.60 & 43.58 & 43.08 \\
XM3600 Text-Image: Swedish & & 66.11 & 59.05 & 60.50 & 58.78 & 50.72 & 51.82 & 55.34 & 47.66 & 47.93 \\
XM3600 Text-Image: Swahili & & 96.30 & 94.01 & 94.73 & 94.55 & 90.09 & 89.57 & 93.85 & 87.47 & 85.67 \\
XM3600 Text-Image: Telugu & & 98.76 & 92.69 & 90.40 & 97.76 & 87.47 & 83.03 & 98.18 & 84.44 & 79.57 \\
XM3600 Text-Image: Thai & & 86.81 & 80.38 & 79.47 & 81.83 & 74.60 & 73.67 & 82.21 & 73.31 & 69.67 \\
XM3600 Text-Image: Turkish & & 72.31 & 65.24 & 65.17 & 65.21 & 55.12 & 56.70 & 62.35 & 53.59 & 52.19 \\
XM3600 Text-Image: Ukrainian & & 75.01 & 66.08 & 65.35 & 68.84 & 57.74 & 55.32 & 66.07 & 54.18 & 50.84 \\
XM3600 Text-Image: Vietnamese & & 70.38 & 64.82 & 64.64 & 61.84 & 54.00 & 53.39 & 58.46 & 50.29 & 48.76 \\
XM3600 Text-Image: Chinese & & 73.98 & 64.78 & 64.96 & 64.87 & 59.03 & 57.33 & 65.25 & 56.15 & 56.68 \\
\hline
Average Western 0-shot Classification & \multirow{4}{*}{Western} & 34.41 & 32.98 & 32.70 & 23.54 & 21.97 & 20.44 & 21.14 & 17.62 & 17.83 \\
Average Western 10-shot Classification & & 30.63 & 30.77 & 30.43 & 23.62 & 23.59 & 23.10 & 22.76 & 21.30 & 21.21 \\
Average Western 0-shot Retrieval & & 48.67 & 45.48 & 46.24 & 44.55 & 39.83 & 39.23 & 41.13 & 36.08 & 35.92 \\
Average Western Classification & & 31.66 & 31.37 & 31.05 & 23.60 & 23.15 & 22.37 & 22.32 & 20.30 & 20.29 \\
\dottedline
Average Dollar Street Classification & \multirow{2}{*}{Culture} & 64.87 & 63.85 & 61.86 & 56.89 & 56.09 & 53.66 & 57.30 & 53.84 & 50.52 \\
Average GeoDE Classification & & 47.23 & 46.85 & 46.39 & 40.72 & 40.60 & 37.01 & 39.16 & 34.24 & 32.35 \\
\dottedline
Average Income Classification & \multirow{3}{*}{Fairness} & 52.03 & 51.87 & 51.59 & 50.22 & 48.09 & 49.02 & 49.99 & 48.57 & 47.35 \\
Average Geographic Classification & & 7.78 & 8.19 & 8.59 & 5.95 & 5.84 & 4.83 & 5.92 & 4.83 & 4.77 \\
Average Demography Classification & & 25.24 & 24.43 & 27.49 & 24.91 & 26.47 & 25.50 & 25.50 & 25.13 & 27.22 \\
\dottedline
Average Multiling: Low-Resource Lang & \multirow{2}{*}{Multiling} & 91.22 & 84.27 & 83.16 & 87.73 & 77.14 & 75.01 & 86.58 & 73.69 & 70.93 \\
Average Multiling: High-Resource Lang & & 63.66 & 55.42 & 55.53 & 55.54 & 46.75 & 45.43 & 53.38 & 43.11 & 41.81 \\
\midrule
\multicolumn{2}{c|}{Average Western} & 36.20 & 35.13 & 35.10 & 29.19 & 27.60 & 26.87 & 27.34 & 24.51 & 24.46 \\
\multicolumn{2}{c|}{Average Culture} & 56.08 & 54.87 & 53.72 & 47.72 & 46.72 & 44.01 & 46.69 & 41.75 & 39.48 \\
\multicolumn{2}{c|}{Average Fairness} & 25.44 & 25.46 & 26.08 & 23.87 & 23.36 & 23.01 & 23.88 & 22.80 & 22.70 \\
\multicolumn{2}{c|}{Average Multiling} & 65.23 & 56.09 & 55.61 & 57.52 & 47.23 & 45.40 & 55.38 & 43.33 & 41.63 \\

\end{longtable}
}
\newpage

\section{Evaluations of Transferability to Generative Models}
\label{appendix:transfer}

The downstream tasks in Table~\ref{tab:transfer_all} are categorized as the following groups and reported in Table~\ref{tab:transfer_avg}:

{\footnotesize

\begin{enumerate}
    \item \textbf{Semantics}: ``COCOcap'', ``NoCaps'', ``COCO-35L (en)'', ``XM3600 (en)'', ``OKVQA'', ``AOKVQA-MC (val)'', ``AOKVQA-DA (val)'', ``GQA'', ``NLVR2'', ``MARVL (avg5)'', ``VizWizVQA (val)'', ``TallyQA (simple)'', ``TallyQA (complex)'', ``CountBenchQA'', ``RefCOCO (testA)'', ``RefCOCO (testB)'', ``RefCOCO+ (testA)'', ``RefCOCO+ (testB)'', ``RefCOCOg (test)''
    \item \textbf{OCR}: ``DocVQA (val)'', ``OCR-VQA'', ``ChartQA (avg)'', ``ChartQA (human)'', ``ChartQA (aug)'', ``SciCap'', ``AI2D'', ``ScienceQA'', ``InfoVQA (val)'', ``TextCaps'', ``TextVQA (val)'', ``ST-VQA (val)'', ``Screen2Words'', ``WidgetCap''
    \item \textbf{Multilinguality}: ``xGQA (avg8)'', ``XM3600 (avg36)'', ``COCO-35L (avg35)''
    \item \textbf{Remote Sensing}: ``RSVQA-lr'', ``RSVQA-hr (test)'', ``RSVQA-hr (test2)''
\end{enumerate}

\begin{longtable}{l|rrr|rrr}
\caption{Detailed evaluation results of the transferability of contrastively trained vision models (ViT-L/16) to generative vision-language models (PaliGemma), with both frozen and unfrozen setups. Task-specific Numbers are reported for vision models trained on 1 billion, 10 billion and 100 billion raw data respectively, using PaliGemma's default fine-tuning configuration.}
\label{tab:transfer_all} \\

\toprule
& \multicolumn{3}{c|}{Frozen ViT} & \multicolumn{3}{c}{Unfrozen ViT} \\
\cmidrule(lr){2-4}\cmidrule(lr){5-7}
Metric & 1B Data & 10B Data & 100B Data & 1B Data & 10B Data & 100B Data \\
\midrule
\endfirsthead  

\endhead  

\endfoot  

\bottomrule
\endlastfoot 

COCOcap & 134.6 & 132.9 & 134.4 & 135.0 & 132.1 & 134.0 \\
NoCaps & 114.1 & 110.5 & 112.8 & 113.4 & 111.4 & 113.3 \\
COCO-35L (avg35) & 107.6 & 105.9 & 108.0 & 107.7 & 106.8 & 107.8 \\
COCO-35L (avg34) & 106.9 & 105.2 & 107.3 & 107.0 & 106.0 & 107.1 \\
COCO-35L (en) & 130.6 & 130.4 & 133.4 & 132.4 & 132.5 & 133.4 \\
XM3600 (en) & 75.5 & 74.9 & 75.2 & 75.3 & 75.4 & 76.0 \\
XM3600 (avg36) & 37.9 & 36.9 & 38.0 & 37.7 & 37.5 & 38.0 \\
Screen2Words & 108.9 & 107.5 & 109.9 & 105.0 & 105.3 & 105.5 \\
TextCaps & 86.5 & 79.3 & 93.2 & 87.6 & 81.8 & 83.8 \\
SciCap & 149.7 & 146.9 & 150.0 & 146.1 & 144.6 & 147.1 \\
WidgetCap & 120.1 & 109.6 & 117.9 & 113.3 & 108.4 & 114.9 \\
VQAv2 (minival) & 79.4 & 78.8 & 79.8 & 79.2 & 78.6 & 78.6 \\
OKVQA & 60.4 & 59.6 & 59.7 & 59.6 & 59.7 & 59.9 \\
AOKVQA-MC (val) & 74.2 & 72.7 & 73.0 & 73.0 & 72.7 & 74.2 \\
AOKVQA-DA (val) & 58.5 & 56.8 & 57.3 & 59.1 & 57.7 & 57.9 \\
GQA & 63.4 & 63.5 & 63.6 & 63.8 & 63.0 & 63.5 \\
NLVR2 & 87.5 & 86.7 & 87.2 & 86.4 & 86.4 & 87.0 \\
MARVL (avg5) & 76.7 & 76.2 & 76.6 & 76.3 & 76.8 & 77.0 \\
AI2D & 69.8 & 70.0 & 70.6 & 68.2 & 68.5 & 68.6 \\
ScienceQA & 95.4 & 94.9 & 94.4 & 94.5 & 92.9 & 94.7 \\
RSVQA-lr & 93.0 & 92.4 & 92.3 & 93.6 & 92.8 & 93.0 \\
RSVQA-hr (test) & 92.5 & 92.5 & 92.7 & 92.6 & 92.6 & 92.6 \\
RSVQA-hr (test2) & 90.4 & 90.4 & 90.5 & 90.5 & 90.4 & 90.6 \\
ChartQA (avg) & 45.1 & 43.6 & 45.0 & 41.4 & 40.3 & 42.5 \\
ChartQA (human) & 31.8 & 31.8 & 32.6 & 29.8 & 28.3 & 30.5 \\
ChartQA (aug) & 58.5 & 55.4 & 57.4 & 53.0 & 52.3 & 54.5 \\
VizWizVQA (val) & 72.3 & 71.2 & 72.8 & 72.0 & 71.6 & 71.9 \\
TallyQA (simple) & 76.6 & 75.7 & 75.9 & 76.6 & 75.7 & 76.9 \\
TallyQA (complex) & 65.0 & 65 & 65.5 & 65.4 & 64.5 & 65.3 \\
CountBenchQA & 68.2 & 69.0 & 67.3 & 60.6 & 61.2 & 63.7 \\
OCR-VQA & 68.3 & 67.5 & 68.2 & 66.9 & 66.0 & 67.1 \\
TextVQA (val) & 44.5 & 41.4 & 44.7 & 41.2 & 40.4 & 41.2 \\
DocVQA (val) & 25.0 & 23.5 & 25.8 & 23.4 & 21.7 & 23.1 \\
InfoVQA (val) & 22.3 & 22.2 & 23 & 21.4 & 22.0 & 22.1 \\
ST-VQA (val) & 46.6 & 42.8 & 46.7 & 43.5 & 40.1 & 43.2 \\
xGQA (avg8) & 55.2 & 55.2 & 55 & 55.6 & 54.5 & 54.8 \\
xGQA (avg7) & 54.1 & 54.0 & 53.8 & 54.5 & 53.3 & 53.6 \\
RefCOCO (testA) & 67.4 & 67.5 & 67.9 & 64.5 & 64.2 & 65.1 \\
RefCOCO (testB) & 62.7 & 62.0 & 63.8 & 60.2 & 59.6 & 60.9 \\
RefCOCO+ (testA) & 63 & 62.7 & 63.5 & 60.2 & 59.9 & 60.3 \\
RefCOCO+ (testB) & 55.6 & 54.9 & 56.2 & 53.2 & 52.5 & 53.3 \\
RefCOCOg (test) & 59.1 & 58.9 & 60 & 56.5 & 56.1 & 57.2 \\
Avg Semantics & 77.1 & 76.4 & 77.2 & 76.0 & 75.4 & 76.4 \\
Avg OCR & 69.5 & 66.9 & 70.0 & 66.8 & 65.2 & 67.0 \\
Avg Multilinguality & 66.9 & 66.0 & 67.0 & 67.0 & 66.3 & 66.9 \\
Avg Remote Sensing & 92.0 & 91.8 & 91.8 & 92.3 & 91.9 & 92.1 \\
Avg & 75.1 & 73.7 & 75.3 & 73.6 & 72.7 & 73.9 \\

\end{longtable}
}

\newpage

\section{Evaluations of Data Quality Filtering}
\label{appendix:quality_filter}

{\footnotesize

\begin{longtable}{l|l|rrrrr}
\caption{Detailed evaluation results of data quality filtering on ViT-L/16 models. All evaluations are conducted on datasets of 5 billion image-text pairs and across different number of seen examples. All metrics are measured by error rate, with the exception of ``Representation Bias'', which is measured by disparity.}
\label{tab:quality_filter_all} \\

\toprule
Metric & Filter & 1B & 5B & 10B & 20B & 30B \\ 
\midrule
\endfirsthead  

\endhead  

\endfoot  

\bottomrule
\endlastfoot 

\multirow[t]{3}{*}{ImageNet 0-shot Classification} & Baseline (en) & 34.67 & 28.17 & 26.68 & 26.15 & 24.32 \\
 & CLIP filtered & 31.18 & 26.76 & 25.14 & 24.39 & 23.90 \\
 & Other filtered & 34.50 & 29.52 & 28.13 & 26.70 & 26.45 \\
\midrule
\multirow[t]{3}{*}{Cifar100 0-shot Classification} & Baseline (en) & 33.05 & 26.08 & 24.37 & 24.52 & 23.99 \\
 & CLIP filtered & 31.69 & 26.96 & 25.37 & 24.68 & 25.76 \\
 & Other filtered & 36.07 & 35.27 & 29.95 & 32.58 & 30.78 \\
\midrule
\multirow[t]{3}{*}{Pet 0-shot Classification} & Baseline (en) & 17.25 & 11.99 & 11.69 & 9.13 & 8.72 \\
 & CLIP filtered & 13.68 & 10.49 & 8.78 & 8.59 & 8.23 \\
 & Other filtered & 14.04 & 9.62 & 8.99 & 7.28 & 6.62 \\
\midrule
\multirow[t]{3}{*}{ImageNet 10-shot Classification} & Baseline (en) & 42.41 & 35.25 & 33.17 & 33.17 & 30.68 \\
 & CLIP filtered & 38.57 & 32.53 & 30.60 & 29.20 & 28.72 \\
 & Other filtered & 38.32 & 32.32 & 30.42 & 29.05 & 28.46 \\
\midrule
\multirow[t]{3}{*}{Cifar100 10-shot Classification} & Baseline (en) & 36.61 & 30.02 & 27.39 & 27.23 & 26.82 \\
 & CLIP filtered & 32.83 & 28.44 & 28.04 & 26.20 & 27.40 \\
 & Other filtered & 35.30 & 35.56 & 31.18 & 32.26 & 31.79 \\
\midrule
\multirow[t]{3}{*}{Pet 10-shot Classification} & Baseline (en) & 22.95 & 16.93 & 15.32 & 15.26 & 11.72 \\
 & CLIP filtered & 17.31 & 11.72 & 10.44 & 8.97 & 8.83 \\
 & Other filtered & 14.15 & 10.38 & 9.08 & 7.63 & 7.52 \\
\midrule
\multirow[t]{3}{*}{Bird 10-shot Classification} & Baseline (en) & 41.18 & 31.69 & 29.91 & 29.60 & 27.37 \\
 & CLIP filtered & 32.38 & 25.20 & 23.85 & 22.21 & 21.95 \\
 & Other filtered & 34.57 & 27.01 & 26.30 & 24.65 & 23.73 \\
\midrule
\multirow[t]{3}{*}{Caltech 10-shot Classification} & Baseline (en) & 10.45 & 9.94 & 9.34 & 9.63 & 9.60 \\
 & CLIP filtered & 11.18 & 10.68 & 10.44 & 10.50 & 10.50 \\
 & Other filtered & 8.97 & 9.25 & 9.01 & 8.30 & 9.06 \\
\midrule
\multirow[t]{3}{*}{Cars 10-shot Classification} & Baseline (en) & 16.47 & 11.03 & 10.16 & 10.05 & 8.94 \\
 & CLIP filtered & 13.07 & 9.70 & 8.89 & 7.75 & 8.01 \\
 & Other filtered & 16.84 & 13.07 & 12.52 & 11.30 & 11.30 \\
\midrule
\multirow[t]{3}{*}{Colorectal Histology 10-shot Classification} & Baseline (en) & 27.80 & 27.17 & 24.77 & 27.03 & 25.33 \\
 & CLIP filtered & 25.97 & 22.90 & 20.80 & 24.23 & 27.13 \\
 & Other filtered & 24.53 & 24.70 & 25.47 & 27.10 & 26.53 \\
\midrule
\multirow[t]{3}{*}{DTD 10-shot Classification} & Baseline (en) & 31.12 & 26.91 & 26.33 & 26.97 & 26.86 \\
 & CLIP filtered & 29.20 & 25.69 & 25.37 & 23.51 & 23.72 \\
 & Other filtered & 28.09 & 26.81 & 24.73 & 24.52 & 23.56 \\
\midrule
\multirow[t]{3}{*}{COCO Image-Text 0-shot Retrieval} & Baseline (en) & 46.80 & 40.28 & 39.30 & 39.18 & 37.04 \\
 & CLIP filtered & 41.06 & 36.04 & 36.48 & 34.84 & 34.02 \\
 & Other filtered & 42.92 & 38.32 & 36.80 & 35.96 & 36.24 \\
\midrule
\multirow[t]{3}{*}{COCO Text-Image 0-shot Retrieval} & Baseline (en) & 62.26 & 56.78 & 54.78 & 55.22 & 53.20 \\
 & CLIP filtered & 59.11 & 55.27 & 54.45 & 53.12 & 53.03 \\
 & Other filtered & 60.53 & 56.01 & 54.60 & 53.23 & 53.27 \\
\midrule
\multirow[t]{3}{*}{Flickr Image-Text 0-shot Retrieval} & Baseline (en) & 16.70 & 11.30 & 11.30 & 11.30 & 10.90 \\
 & CLIP filtered & 14.80 & 9.90 & 9.70 & 9.60 & 8.90 \\
 & Other filtered & 16.70 & 13.80 & 12.60 & 13.10 & 12.00 \\
\midrule
\multirow[t]{3}{*}{Flickr Text-Image 0-shot Retrieval} & Baseline (en) & 32.26 & 24.78 & 24.74 & 24.90 & 22.66 \\
 & CLIP filtered & 29.52 & 24.98 & 23.34 & 22.12 & 22.02 \\
 & Other filtered & 32.84 & 27.18 & 26.48 & 24.82 & 24.32 \\
\midrule
\multirow[t]{3}{*}{Dollar Street 0-shot Classification} & Baseline (en) & 54.67 & 50.44 & 49.81 & 49.98 & 49.37 \\
 & CLIP filtered & 53.71 & 52.58 & 51.88 & 50.63 & 51.44 \\
 & Other filtered & 50.23 & 47.63 & 47.86 & 47.45 & 47.08 \\
\midrule
\multirow[t]{3}{*}{Dollar Street 10-shot Classification} & Baseline (en) & 84.87 & 79.27 & 77.18 & 76.21 & 72.54 \\
 & CLIP filtered & 88.86 & 84.59 & 84.73 & 82.80 & 82.80 \\
 & Other filtered & 90.16 & 89.46 & 87.91 & 88.72 & 87.77 \\
\midrule
\multirow[t]{3}{*}{GeoDE 0-shot Classification} & Baseline (en) & 8.98 & 6.48 & 6.43 & 6.26 & 6.23 \\
 & CLIP filtered & 9.64 & 8.54 & 8.02 & 7.42 & 7.22 \\
 & Other filtered & 9.50 & 7.69 & 7.50 & 7.50 & 7.53 \\
\midrule
\multirow[t]{3}{*}{GeoDE (country) 10-shot Classification} & Baseline (en) & 84.29 & 77.28 & 73.22 & 73.37 & 68.85 \\
 & CLIP filtered & 85.82 & 81.98 & 80.11 & 78.08 & 78.24 \\
 & Other filtered & 91.37 & 89.52 & 88.30 & 87.65 & 86.76 \\
\midrule
\multirow[t]{3}{*}{GeoDE (region) 10-shot Classification} & Baseline (en) & 66.67 & 61.66 & 57.71 & 58.77 & 55.78 \\
 & CLIP filtered & 70.68 & 68.16 & 66.99 & 64.81 & 63.68 \\
 & Other filtered & 75.82 & 72.39 & 72.95 & 72.13 & 71.27 \\
\midrule
\multirow[t]{3}{*}{GLDv2 0-shot Classification} & Baseline (en) & 65.50 & 53.18 & 50.13 & 49.48 & 44.16 \\
 & CLIP filtered & 61.15 & 52.46 & 49.55 & 47.41 & 46.37 \\
 & Other filtered & 80.87 & 74.06 & 72.37 & 72.37 & 70.17 \\
\midrule
\multirow[t]{3}{*}{Representation Bias} & Baseline (en) & 33.89 & 28.22 & 36.00 & 33.52 & 30.96 \\
 & CLIP filtered & 11.46 & 19.14 & 20.03 & 26.57 & 14.05 \\
 & Other filtered & 39.31 & 36.44 & 39.01 & 40.57 & 35.51 \\
\midrule
\multirow[t]{3}{*}{Income 0-200 Classification} & Baseline (en) & 71.31 & 67.22 & 68.34 & 67.50 & 67.04 \\
 & CLIP filtered & 69.36 & 69.36 & 68.71 & 66.67 & 67.87 \\
 & Other filtered & 69.36 & 67.97 & 65.65 & 66.11 & 66.67 \\
\midrule
\multirow[t]{3}{*}{Income 200-285 Classification} & Baseline (en) & 60.15 & 55.33 & 54.87 & 54.49 & 55.33 \\
 & CLIP filtered & 58.48 & 57.46 & 56.63 & 54.59 & 56.63 \\
 & Other filtered & 54.22 & 50.88 & 52.64 & 51.16 & 51.16 \\
\midrule
\multirow[t]{3}{*}{Income 285-685 Classification} & Baseline (en) & 46.61 & 42.99 & 41.04 & 42.43 & 40.76 \\
 & CLIP filtered & 46.43 & 44.75 & 44.20 & 42.90 & 43.45 \\
 & Other filtered & 40.95 & 39.09 & 39.37 & 39.37 & 37.70 \\
\midrule
\multirow[t]{3}{*}{Income \textgreater1998 Classification} & Baseline (en) & 40.56 & 36.19 & 34.98 & 35.44 & 34.33 \\
 & CLIP filtered & 40.56 & 38.70 & 37.95 & 38.33 & 37.77 \\
 & Other filtered & 36.37 & 32.56 & 33.77 & 33.12 & 32.74 \\
\midrule
\multirow[t]{3}{*}{Africa} & Baseline (en) & 11.51 & 8.19 & 7.88 & 7.72 & 7.85 \\
 & CLIP filtered & 11.00 & 9.74 & 9.37 & 9.28 & 8.44 \\
 & Other filtered & 12.04 & 9.97 & 9.51 & 9.85 & 9.88 \\
\midrule
\multirow[t]{3}{*}{Americas} & Baseline (en) & 8.59 & 6.74 & 6.15 & 6.37 & 6.27 \\
 & CLIP filtered & 9.57 & 8.60 & 8.30 & 7.29 & 7.16 \\
 & Other filtered & 9.63 & 7.68 & 7.32 & 7.53 & 7.48 \\
\midrule
\multirow[t]{3}{*}{EastAsia} & Baseline (en) & 9.90 & 7.10 & 7.37 & 7.29 & 6.71 \\
 & CLIP filtered & 10.45 & 9.34 & 8.88 & 7.72 & 7.67 \\
 & Other filtered & 10.52 & 8.92 & 8.63 & 8.21 & 8.48 \\
\midrule
\multirow[t]{3}{*}{Europe} & Baseline (en) & 6.75 & 4.82 & 5.29 & 5.01 & 5.17 \\
 & CLIP filtered & 7.71 & 6.89 & 6.52 & 5.52 & 6.01 \\
 & Other filtered & 7.29 & 5.62 & 5.57 & 5.45 & 5.51 \\
\midrule
\multirow[t]{3}{*}{SouthEastAsia} & Baseline (en) & 8.69 & 6.23 & 6.00 & 5.77 & 6.01 \\
 & CLIP filtered & 9.74 & 8.47 & 7.40 & 7.74 & 7.32 \\
 & Other filtered & 8.89 & 7.28 & 7.47 & 7.16 & 7.11 \\
\midrule
\multirow[t]{3}{*}{WestAsia} & Baseline (en) & 8.14 & 5.61 & 5.64 & 5.17 & 5.08 \\
 & CLIP filtered & 9.24 & 8.16 & 7.59 & 6.75 & 6.52 \\
 & Other filtered & 8.32 & 6.34 & 6.17 & 6.47 & 6.35 \\
\midrule
\multirow[t]{3}{*}{Perceived Gender} & Baseline (en) & 8.41 & 6.42 & 5.78 & 5.98 & 5.64 \\
 & CLIP filtered & 8.43 & 7.63 & 8.08 & 7.56 & 6.35 \\
 & Other filtered & 13.08 & 10.64 & 11.13 & 11.02 & 9.55 \\
\midrule
\multirow[t]{3}{*}{Perceived Race} & Baseline (en) & 37.87 & 44.74 & 43.93 & 48.30 & 44.95 \\
 & CLIP filtered & 33.08 & 40.63 & 38.98 & 41.89 & 43.04 \\
 & Other filtered & 53.52 & 52.46 & 53.21 & 52.83 & 56.43 \\
\midrule
\multirow[t]{3}{*}{Average Western 0-shot Classification} & Baseline (en) & 28.33 & 22.08 & 20.91 & 19.93 & 19.01 \\
 & CLIP filtered & 25.52 & 21.40 & 19.76 & 19.22 & 19.30 \\
 & Other filtered & 28.20 & 24.81 & 22.36 & 22.18 & 21.28 \\
\midrule
\multirow[t]{3}{*}{Average Western 10-shot Classification} & Baseline (en) & 28.62 & 23.62 & 22.05 & 22.37 & 20.92 \\
 & CLIP filtered & 25.06 & 20.86 & 19.80 & 19.07 & 19.53 \\
 & Other filtered & 25.10 & 22.39 & 21.09 & 20.60 & 20.25 \\
\midrule
\multirow[t]{3}{*}{Average Western 0-shot Retrieval} & Baseline (en) & 39.50 & 33.29 & 32.53 & 32.65 & 30.95 \\
 & CLIP filtered & 36.12 & 31.55 & 30.99 & 29.92 & 29.49 \\
 & Other filtered & 38.25 & 33.83 & 32.62 & 31.78 & 31.46 \\
\midrule
\multirow[t]{3}{*}{Average Western Classification} & Baseline (en) & 28.54 & 23.20 & 21.74 & 21.70 & 20.40 \\
 & CLIP filtered & 25.19 & 21.01 & 19.79 & 19.11 & 19.47 \\
 & Other filtered & 25.94 & 23.05 & 21.43 & 21.03 & 20.53 \\
\midrule
\multirow[t]{3}{*}{Average Dollar Street Classification} & Baseline (en) & 69.77 & 64.86 & 63.50 & 63.09 & 60.96 \\
 & CLIP filtered & 71.29 & 68.58 & 68.30 & 66.71 & 67.12 \\
 & Other filtered & 70.19 & 68.55 & 67.89 & 68.08 & 67.42 \\
\midrule
\multirow[t]{3}{*}{Average GeoDE Classification} & Baseline (en) & 53.32 & 48.48 & 45.79 & 46.13 & 43.62 \\
 & CLIP filtered & 55.38 & 52.89 & 51.71 & 50.10 & 49.71 \\
 & Other filtered & 58.90 & 56.54 & 56.25 & 55.76 & 55.18 \\
\midrule
\multirow[t]{3}{*}{Average Income Classification} & Baseline (en) & 54.66 & 50.43 & 49.81 & 49.97 & 49.36 \\
 & CLIP filtered & 53.71 & 52.57 & 51.87 & 50.62 & 51.43 \\
 & Other filtered & 50.22 & 47.62 & 47.86 & 47.44 & 47.07 \\
\midrule
\multirow[t]{3}{*}{Average Geographic Classification} & Baseline (en) & 8.93 & 6.44 & 6.39 & 6.22 & 6.18 \\
 & CLIP filtered & 9.62 & 8.53 & 8.01 & 7.39 & 7.19 \\
 & Other filtered & 9.45 & 7.63 & 7.44 & 7.45 & 7.47 \\
\midrule
\multirow[t]{3}{*}{Average Demography Classification} & Baseline (en) & 23.14 & 25.58 & 24.86 & 27.14 & 25.30 \\
 & CLIP filtered & 20.76 & 24.13 & 23.53 & 24.72 & 24.70 \\
 & Other filtered & 33.30 & 31.55 & 32.17 & 31.93 & 32.99 \\
\midrule
\multirow[t]{3}{*}{Average Western} & Baseline (en) & 31.47 & 25.89 & 24.62 & 24.62 & 23.21 \\
 & CLIP filtered & 28.10 & 23.82 & 22.78 & 21.99 & 22.14 \\
 & Other filtered & 29.22 & 25.92 & 24.42 & 23.90 & 23.44 \\
\midrule
\multirow[t]{3}{*}{Average Culture Diversity} & Baseline (en) & 60.83 & 54.72 & 52.41 & 52.34 & 49.49 \\
 & CLIP filtered & 61.64 & 58.05 & 56.88 & 55.19 & 54.96 \\
 & Other filtered & 66.33 & 63.46 & 62.82 & 62.64 & 61.76 \\
\midrule
\multirow[t]{3}{*}{Average Fairness} & Baseline (en) & 26.54 & 24.30 & 23.94 & 24.29 & 23.76 \\
 & CLIP filtered & 26.17 & 25.81 & 25.22 & 24.69 & 24.85 \\
 & Other filtered & 27.02 & 24.95 & 25.04 & 24.86 & 24.92 \\

\end{longtable}
}

\newpage

\section{Evaluations of Language Rebalancing}
\label{appendix:lang_rebalance}

\FloatBarrier

{\footnotesize

\begin{longtable}{l|rr|rr|rr} 
\caption{Detailed evaluation results of the rebalancing of low-resource languages on ViT-L/16 models and datasets of 1/10/100 billion scales, with 100 billion examples seen in training. All metrics are measured by error rate, with the exception of ``Representation Bias'', which is measured by disparity.}
\label{tab:lang_rebalance_all} \\

\toprule
& \multicolumn{2}{c}{1B Data} & \multicolumn{2}{|c|}{10B Data} & \multicolumn{2}{c}{100B Data} \\
\cmidrule(lr){2-3}\cmidrule(lr){4-5}\cmidrule(lr){6-7}
Metric & Before & After & Before & After & Before & After \\ 
\midrule
\endfirsthead  

\endhead  

\endfoot  

\bottomrule
\endlastfoot 

ImageNet 0-shot Classification & 31.23 & 31.39 & 29.70 & 30.47 & 28.49 & 28.80 \\
Cifar100 0-shot Classification & 25.02 & 24.96 & 23.75 & 24.04 & 23.36 & 23.51 \\
Pet 0-shot Classification & 14.36 & 13.00 & 12.46 & 12.05 & 9.46 & 11.23 \\
ImageNet 10-shot Classification & 35.11 & 34.94 & 34.95 & 34.99 & 33.71 & 33.89 \\
Cifar100 10-shot Classification & 27.50 & 27.82 & 26.70 & 26.50 & 25.49 & 25.05 \\
Pet 10-shot Classification & 12.32 & 13.71 & 12.48 & 15.59 & 11.80 & 13.46 \\
Bird 10-shot Classification & 44.05 & 42.75 & 45.25 & 45.29 & 44.29 & 42.89 \\
Caltech 10-shot Classification & 6.41 & 8.09 & 7.40 & 8.97 & 7.53 & 8.35 \\
Cars 10-shot Classification & 11.14 & 11.34 & 11.33 & 11.54 & 11.47 & 11.21 \\
Colorectal Histology 10-shot Classification & 24.00 & 25.50 & 23.53 & 24.43 & 22.57 & 28.00 \\
DTD 10-shot Classification & 28.46 & 29.31 & 27.07 & 27.39 & 27.93 & 29.04 \\
COCO Image-Text 0-shot Retrieval & 49.70 & 52.92 & 47.18 & 50.28 & 45.28 & 45.90 \\
COCO Text-Image 0-shot Retrieval & 68.16 & 67.50 & 64.32 & 63.60 & 62.51 & 62.16 \\
Flickr Image-Text 0-shot Retrieval & 20.40 & 24.30 & 15.50 & 20.30 & 16.60 & 16.40 \\
Flickr Text-Image 0-shot Retrieval & 39.94 & 37.88 & 32.32 & 32.64 & 32.52 & 33.30 \\
Dollar Street 0-shot Classification & 50.23 & 51.16 & 48.10 & 49.42 & 49.03 & 49.23 \\
Dollar Street 10-shot Classification & 63.56 & 65.04 & 64.09 & 65.51 & 58.29 & 59.42 \\
GeoDE 0-shot Classification & 6.01 & 6.03 & 5.90 & 5.97 & 4.88 & 5.42 \\
GeoDE (country) 10-shot Classification & 61.94 & 59.79 & 62.31 & 60.52 & 57.85 & 53.34 \\
GeoDE (region) 10-shot Classification & 54.21 & 53.99 & 53.59 & 53.30 & 48.29 & 48.05 \\
GLDv2 0-shot Classification & 50.39 & 51.82 & 46.37 & 47.73 & 45.72 & 44.29 \\
Representation Bias & 38.18 & 35.21 & 36.35 & 32.61 & 35.51 & 32.74 \\
Income 0-200 Classification & 66.30 & 67.32 & 64.35 & 65.83 & 66.30 & 65.37 \\
Income 200-285 Classification & 55.33 & 54.22 & 52.18 & 53.48 & 53.38 & 53.20 \\
Income 285-685 Classification & 42.71 & 44.75 & 41.32 & 42.80 & 40.48 & 40.76 \\
Income \textgreater1998 Classification & 36.56 & 38.33 & 34.51 & 35.53 & 35.91 & 37.58 \\
Africa & 7.99 & 8.34 & 8.24 & 7.81 & 6.55 & 7.46 \\
Americas & 6.03 & 5.51 & 5.57 & 5.84 & 4.92 & 5.20 \\
EastAsia & 5.98 & 6.07 & 5.96 & 5.90 & 4.56 & 5.27 \\
Europe & 4.81 & 4.41 & 4.20 & 4.23 & 3.75 & 4.00 \\
SouthEastAsia & 5.78 & 6.21 & 5.78 & 6.15 & 5.02 & 5.50 \\
WestAsia & 5.11 & 5.30 & 5.30 & 5.67 & 4.19 & 4.79 \\
Perceived Gender & 5.25 & 5.27 & 6.06 & 5.96 & 4.97 & 5.03 \\
Perceived Race & 44.57 & 49.02 & 46.88 & 45.89 & 46.04 & 47.35 \\
Crossmodal-3600 Image-Text Retrieval: Arabic & 53.58 & 56.44 & 45.00 & 45.89 & 44.56 & 44.78 \\
Crossmodal-3600 Image-Text Retrieval: Bengali & 90.81 & 76.03 & 66.36 & 63.53 & 63.75 & 61.47 \\
Crossmodal-3600 Image-Text Retrieval: Czech & 52.31 & 52.81 & 43.81 & 43.36 & 42.22 & 41.61 \\
Crossmodal-3600 Image-Text Retrieval: Danish & 45.08 & 45.22 & 35.06 & 34.81 & 31.00 & 32.53 \\
Crossmodal-3600 Image-Text Retrieval: German & 30.61 & 32.00 & 24.28 & 24.36 & 24.03 & 23.11 \\
Crossmodal-3600 Image-Text Retrieval: Greek & 67.86 & 70.17 & 53.64 & 53.42 & 50.14 & 51.94 \\
Crossmodal-3600 Image-Text Retrieval: English & 54.14 & 54.58 & 52.42 & 51.58 & 51.67 & 50.89 \\
Crossmodal-3600 Image-Text Retrieval: Spanish & 41.56 & 43.50 & 38.44 & 38.00 & 35.81 & 35.89 \\
Crossmodal-3600 Image-Text Retrieval: Persian & 49.64 & 55.33 & 38.97 & 41.97 & 40.17 & 38.11 \\
Crossmodal-3600 Image-Text Retrieval: Finnish & 59.25 & 60.11 & 42.67 & 42.42 & 39.06 & 40.28 \\
Crossmodal-3600 Image-Text Retrieval: Filipino & 82.72 & 72.56 & 72.86 & 62.72 & 71.36 & 60.22 \\
Crossmodal-3600 Image-Text Retrieval: French & 39.08 & 39.72 & 31.78 & 31.47 & 29.92 & 29.61 \\
Crossmodal-3600 Image-Text Retrieval: Hindi & 77.67 & 71.67 & 65.67 & 65.44 & 63.47 & 63.53 \\
Crossmodal-3600 Image-Text Retrieval: Croatian & 53.08 & 53.72 & 37.94 & 38.86 & 35.78 & 35.64 \\
Crossmodal-3600 Image-Text Retrieval: Hungarian & 53.81 & 54.61 & 38.64 & 37.81 & 34.42 & 34.78 \\
Crossmodal-3600 Image-Text Retrieval: Indonesian & 35.83 & 37.47 & 28.47 & 30.94 & 28.53 & 28.42 \\
Crossmodal-3600 Image-Text Retrieval: Italian & 38.42 & 40.69 & 33.33 & 33.50 & 30.97 & 31.03 \\
Crossmodal-3600 Image-Text Retrieval: Hebrew & 56.75 & 47.75 & 39.44 & 37.39 & 35.72 & 34.19 \\
Crossmodal-3600 Image-Text Retrieval: Japanese & 59.00 & 61.58 & 45.42 & 45.78 & 44.97 & 46.69 \\
Crossmodal-3600 Image-Text Retrieval: Korean & 50.75 & 53.06 & 40.33 & 40.00 & 38.31 & 38.58 \\
Crossmodal-3600 Image-Text Retrieval: Maori & 99.58 & 97.94 & 99.22 & 95.00 & 99.25 & 96.08 \\
Crossmodal-3600 Image-Text Retrieval: Dutch & 47.11 & 48.06 & 41.14 & 41.42 & 38.39 & 39.94 \\
Crossmodal-3600 Image-Text Retrieval: Norwegian & 45.33 & 46.81 & 36.11 & 36.72 & 34.28 & 34.47 \\
Crossmodal-3600 Image-Text Retrieval: Polish & 45.97 & 45.81 & 35.50 & 35.61 & 34.11 & 34.33 \\
Crossmodal-3600 Image-Text Retrieval: Portuguese & 43.33 & 42.53 & 36.03 & 38.33 & 34.56 & 34.11 \\
Crossmodal-3600 Image-Text Retrieval: Quechua & 94.64 & 94.97 & 93.53 & 93.83 & 93.92 & 93.42 \\
Crossmodal-3600 Image-Text Retrieval: Romanian & 52.19 & 52.72 & 38.31 & 38.06 & 35.39 & 34.86 \\
Crossmodal-3600 Image-Text Retrieval: Russian & 42.78 & 45.00 & 35.14 & 35.11 & 33.22 & 33.42 \\
Crossmodal-3600 Image-Text Retrieval: Swedish & 44.50 & 46.19 & 34.94 & 36.06 & 34.78 & 34.19 \\
Crossmodal-3600 Image-Text Retrieval: Swahili & 89.94 & 75.06 & 81.33 & 67.64 & 79.47 & 65.81 \\
Crossmodal-3600 Image-Text Retrieval: Telugu & 96.08 & 81.00 & 76.67 & 67.78 & 69.69 & 66.33 \\
Crossmodal-3600 Image-Text Retrieval: Thai & 72.61 & 74.72 & 59.47 & 60.50 & 58.86 & 59.92 \\
Crossmodal-3600 Image-Text Retrieval: Turkish & 52.78 & 54.94 & 40.72 & 41.25 & 39.72 & 39.89 \\
Crossmodal-3600 Image-Text Retrieval: Ukrainian & 55.19 & 57.33 & 41.25 & 40.97 & 37.83 & 39.19 \\
Crossmodal-3600 Image-Text Retrieval: Vietnamese & 43.19 & 42.22 & 34.00 & 35.22 & 32.44 & 32.86 \\
Crossmodal-3600 Image-Text Retrieval: Chinese & 53.67 & 54.81 & 42.47 & 44.67 & 42.50 & 43.97 \\
Crossmodal-3600 Text-Image Retrieval: Arabic & 67.49 & 65.43 & 59.74 & 59.02 & 59.86 & 59.70 \\
Crossmodal-3600 Text-Image Retrieval: Bengali & 95.17 & 83.83 & 79.72 & 75.56 & 77.31 & 73.33 \\
Crossmodal-3600 Text-Image Retrieval: Czech & 65.52 & 65.19 & 58.57 & 59.19 & 58.18 & 57.56 \\
Crossmodal-3600 Text-Image Retrieval: Danish & 60.01 & 59.93 & 51.18 & 52.77 & 49.50 & 49.74 \\
Crossmodal-3600 Text-Image Retrieval: German & 45.85 & 47.48 & 39.88 & 40.72 & 39.75 & 39.50 \\
Crossmodal-3600 Text-Image Retrieval: Greek & 77.96 & 75.46 & 69.11 & 69.24 & 67.35 & 68.25 \\
Crossmodal-3600 Text-Image Retrieval: English & 58.97 & 56.93 & 57.57 & 57.52 & 56.32 & 56.51 \\
Crossmodal-3600 Text-Image Retrieval: Spanish & 52.64 & 52.79 & 49.06 & 49.90 & 48.31 & 48.76 \\
Crossmodal-3600 Text-Image Retrieval: Persian & 62.65 & 63.27 & 55.06 & 55.54 & 56.09 & 54.64 \\
Crossmodal-3600 Text-Image Retrieval: Finnish & 72.96 & 72.06 & 59.11 & 58.61 & 56.24 & 56.42 \\
Crossmodal-3600 Text-Image Retrieval: Filipino & 90.89 & 83.32 & 83.98 & 78.41 & 83.70 & 74.94 \\
Crossmodal-3600 Text-Image Retrieval: French & 48.33 & 49.81 & 43.31 & 44.62 & 42.10 & 42.34 \\
Crossmodal-3600 Text-Image Retrieval: Hindi & 87.43 & 83.45 & 81.38 & 80.96 & 80.01 & 79.22 \\
Crossmodal-3600 Text-Image Retrieval: Croatian & 66.68 & 65.73 & 54.42 & 56.10 & 54.22 & 53.60 \\
Crossmodal-3600 Text-Image Retrieval: Hungarian & 66.49 & 66.66 & 53.73 & 54.57 & 50.75 & 51.16 \\
Crossmodal-3600 Text-Image Retrieval: Indonesian & 50.28 & 49.62 & 44.05 & 44.58 & 43.97 & 44.30 \\
Crossmodal-3600 Text-Image Retrieval: Italian & 47.96 & 49.51 & 42.80 & 45.41 & 42.60 & 42.66 \\
Crossmodal-3600 Text-Image Retrieval: Hebrew & 69.11 & 60.25 & 56.25 & 55.62 & 54.14 & 51.65 \\
Crossmodal-3600 Text-Image Retrieval: Japanese & 69.56 & 71.62 & 62.34 & 63.34 & 58.44 & 61.42 \\
Crossmodal-3600 Text-Image Retrieval: Korean & 64.52 & 64.72 & 56.76 & 57.83 & 56.51 & 57.58 \\
Crossmodal-3600 Text-Image Retrieval: Maori & 99.73 & 97.92 & 99.56 & 96.30 & 99.62 & 96.19 \\
Crossmodal-3600 Text-Image Retrieval: Dutch & 57.41 & 58.78 & 52.02 & 53.88 & 51.48 & 51.82 \\
Crossmodal-3600 Text-Image Retrieval: Norwegian & 61.54 & 61.46 & 53.81 & 54.35 & 52.99 & 53.50 \\
Crossmodal-3600 Text-Image Retrieval: Polish & 56.06 & 56.43 & 47.92 & 49.96 & 47.09 & 47.16 \\
Crossmodal-3600 Text-Image Retrieval: Portuguese & 54.54 & 54.07 & 49.48 & 51.03 & 48.72 & 48.34 \\
Crossmodal-3600 Text-Image Retrieval: Quechua & 97.88 & 97.89 & 98.14 & 98.03 & 98.04 & 97.88 \\
Crossmodal-3600 Text-Image Retrieval: Romanian & 65.20 & 65.55 & 54.05 & 54.79 & 52.41 & 51.93 \\
Crossmodal-3600 Text-Image Retrieval: Russian & 53.47 & 53.75 & 47.58 & 48.43 & 45.36 & 46.83 \\
Crossmodal-3600 Text-Image Retrieval: Swedish & 58.78 & 59.12 & 50.72 & 52.50 & 51.82 & 50.97 \\
Crossmodal-3600 Text-Image Retrieval: Swahili & 94.55 & 84.91 & 90.09 & 80.20 & 89.57 & 78.20 \\
Crossmodal-3600 Text-Image Retrieval: Telugu & 97.76 & 87.85 & 87.47 & 82.04 & 83.03 & 80.15 \\
Crossmodal-3600 Text-Image Retrieval: Thai & 81.83 & 80.83 & 74.60 & 75.72 & 73.67 & 75.03 \\
Crossmodal-3600 Text-Image Retrieval: Turkish & 65.21 & 64.41 & 55.12 & 58.01 & 56.70 & 56.82 \\
Crossmodal-3600 Text-Image Retrieval: Ukrainian & 68.84 & 68.01 & 57.74 & 59.49 & 55.32 & 57.30 \\
Crossmodal-3600 Text-Image Retrieval: Vietnamese & 61.84 & 61.28 & 54.00 & 55.01 & 53.39 & 53.51 \\
Crossmodal-3600 Text-Image Retrieval: Chinese & 64.87 & 65.56 & 59.03 & 61.21 & 57.33 & 59.49 \\
Average Western 0-shot Classification & 23.54 & 23.12 & 21.97 & 22.18 & 20.44 & 21.18 \\
Average Western 10-shot Classification & 23.62 & 24.18 & 23.59 & 24.34 & 23.10 & 23.99 \\
Average Western 0-shot Retrieval & 44.55 & 45.65 & 39.83 & 41.70 & 39.23 & 39.44 \\
Average Western Classification & 23.60 & 23.89 & 23.15 & 23.75 & 22.37 & 23.22 \\
Average Dollar Street Classification & 56.89 & 58.10 & 56.09 & 57.46 & 53.66 & 54.33 \\
Average GeoDE Classification & 40.72 & 39.94 & 40.60 & 39.93 & 37.01 & 35.60 \\
Average Income Classification & 50.22 & 51.15 & 48.09 & 49.41 & 49.02 & 49.23 \\
Average Geographic Classification & 5.95 & 5.97 & 5.84 & 5.93 & 4.83 & 5.37 \\
Average Demography Classification & 24.91 & 27.14 & 26.47 & 25.93 & 25.50 & 26.19 \\
Average Multilingual: Low-Resource Lang & 87.73 & 78.82 & 77.14 & 72.04 & 75.01 & 70.10 \\
Average Multilingual: High-Resource Lang & 55.54 & 56.21 & 46.75 & 47.53 & 45.43 & 45.75 \\
Average Western & 29.19 & 29.69 & 27.60 & 28.54 & 26.87 & 27.55 \\
Average Culture Diversity & 47.72 & 47.97 & 46.72 & 47.07 & 44.01 & 43.29 \\
Average Fairness & 23.87 & 24.56 & 23.36 & 23.76 & 23.01 & 23.46 \\
Average Multilingual & 57.52 & 56.64 & 47.23 & 46.43 & 45.40 & 44.61 \\

\end{longtable}
}
\section{Distribution of Languages}
\label{appendix:lang_distribution}

We reuse the 35 languages\footnote{``Quechua'' is excluded as it is not supported by the language detection method we used.} reported in Crossmodal-3600 benchmark~\citep{thapliyal2022crossmodal} for multilingual experiments.

{\footnotesize

\begin{longtable}{l|l|r}
\caption{Distribution of the 35 languages used in multilingual evaluations.}
\label{tab:lang_distribution} \\

\toprule
Language & Type & Pages (\%) \\
\midrule
\endfirsthead  

\toprule
Language & Type & Percentage (\%) \\
\midrule
\endhead  

\bottomrule
\endfoot  

\bottomrule
\endlastfoot 

Maori & Low-resource & 0.001 \\
Telugu & Low-resource & 0.036 \\
Swahili & Low-resource & 0.046 \\
Filipino & Low-resource & 0.111 \\
Bengali & Low-resource & 0.113 \\
Hebrew & Low-resource & 0.240 \\
Hindi & Low-resource & 0.267 \\
Croatian & High-resource & 0.284 \\
Norwegian & High-resource & 0.290 \\
Finnish & High-resource & 0.296 \\
Danish & High-resource & 0.370 \\
Hungarian & High-resource & 0.378 \\
Ukrainian & High-resource & 0.476 \\
Romanian & High-resource & 0.489 \\
Greek & High-resource & 0.560 \\
Swedish & High-resource & 0.660 \\
Czech & High-resource & 0.727 \\
Persian & High-resource & 0.881 \\
Thai & High-resource & 1.167 \\
Dutch & High-resource & 1.173 \\
Arabic & High-resource & 1.258 \\
Vietnamese & High-resource & 1.337 \\
Turkish & High-resource & 1.554 \\
Polish & High-resource & 1.825 \\
Italian & High-resource & 1.964 \\
Korean & High-resource & 2.519 \\
Portuguese & High-resource & 3.054 \\
Indonesian & High-resource & 3.181 \\
French & High-resource & 3.354 \\
Chinese & High-resource & 3.544 \\
German & High-resource & 3.869 \\
Russian & High-resource & 6.981 \\
Spanish & High-resource & 8.214 \\
Japanese & High-resource & 8.752 \\
English & High-resource & 35.353 \\
\textbf{Low-resource All} & Low-resource & 0.814 \\
\textbf{High-resource All} & High-resource & 94.510 \\

\end{longtable}
}

\begin{figure}[h]
    \centering
    \includegraphics[width=0.9\linewidth]{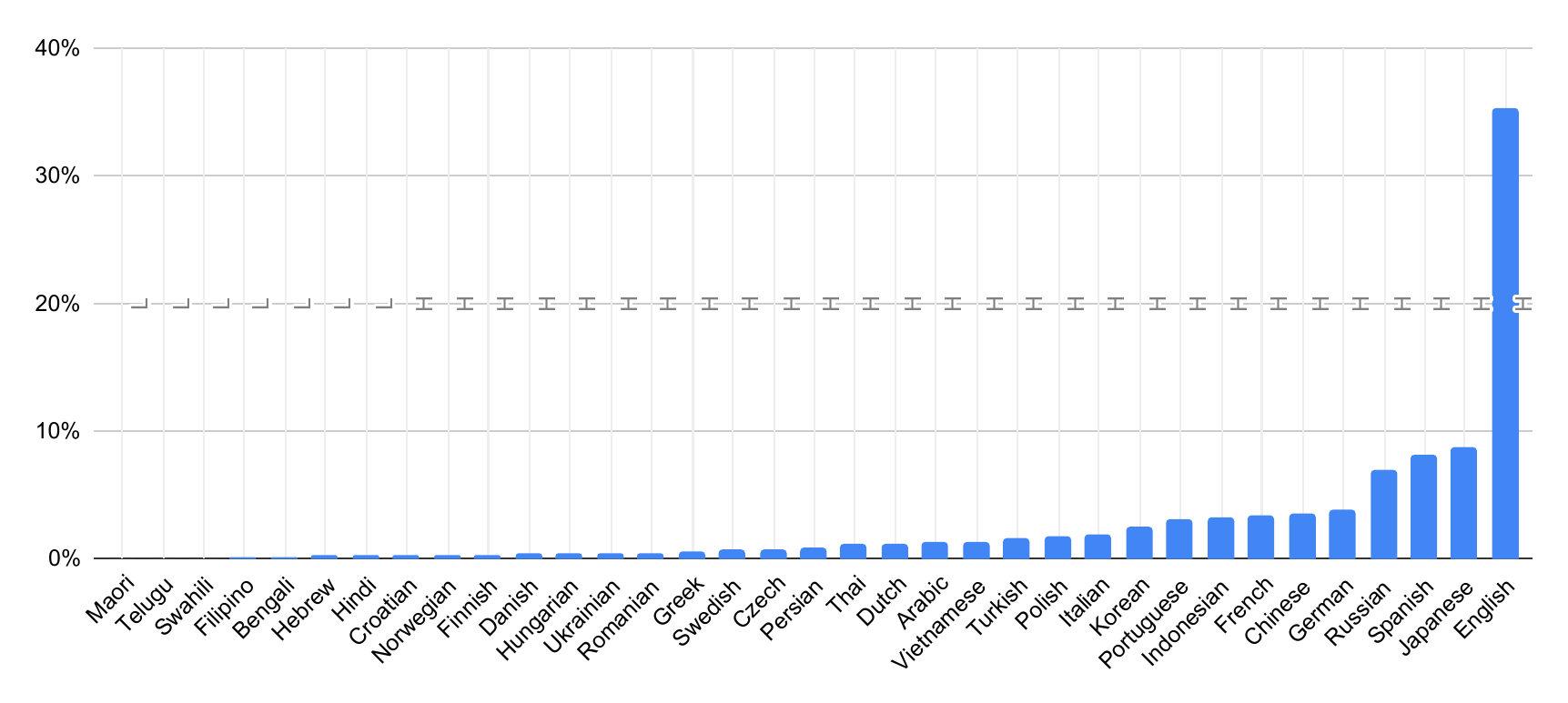}
    \caption{Visualization of the language distribution, where ``L'' and ``H'' denote low-resource and high-resource language respectively.}
    \label{fig:lang_distribution}
\end{figure}
\FloatBarrier

\section{Computation-based Scaling Law}
\label{appendix:compute_scaling_law}

\begin{table}
\centering
\caption{ImageNet Zero-shot Classification.}
\begin{tabular}{ll|rrrr}
\toprule
Model & Data & 3B & 10B & 33B & 100B \\
\midrule
B & 100B & 52.6 & 56.8 & 59.9 & 61 \\
B & 10B & 52.1 & 56.2 & 59.3 & 60.6 \\
B & 1B & 51.5 & 55.6 & 58 & 58.8 \\
H & 100B & 66 & 70.5 & 73.4 & 75.1 \\
H & 10B & 65.5 & 70.4 & 73 & 74.4 \\
H & 1B & 64.3 & 68.5 & 71 & 70.4 \\
L & 100B & 61.6 & 66.2 & 69.3 & 71.5 \\
L & 10B & 61.7 & 66.1 & 69.1 & 70.3 \\
L & 1B & 61.2 & 65.3 & 67.9 & 68.8 \\
\bottomrule
\end{tabular}
\end{table}

\begin{table}
\centering
\caption{COCO Image-To-Text Zero-shot Retrieval.}
\begin{tabular}{ll|rrrr}
\toprule
Model & Data & 3B & 10B & 33B & 100B \\
\midrule
B & 100B & 41.2 & 44 & 45.7 & 46.6 \\ 
B & 10B & 41.3 & 45.1 & 47 & 48.4 \\ 
B & 1B & 40.2 & 43 & 44.7 & 43.5 \\ 
H & 100B & 49.8 & 52.1 & 56.1 & 57.5 \\ 
H & 10B & 49.8 & 53.1 & 55.8 & 58 \\ 
H & 1B & 47.9 & 52 & 52 & 51.4 \\ 
L & 100B & 48.4 & 50.2 & 52.5 & 54.7 \\ 
L & 10B & 47 & 49.8 & 52.3 & 52.8 \\ 
L & 1B & 46.4 & 48.9 & 50.6 & 50.3 \\ 
\bottomrule
\end{tabular}
\end{table}

\begin{table}
\centering
\caption{DollarStreet Geoloc 10-shot Retrieval.}
\begin{tabular}{ll|rrrr}
\toprule
Model & Data & 3B & 10B & 33B & 100B \\
\midrule
B & 100B & 21.7 & 24.7 & 26 & 27.9 \\ 
B & 10B & 18.3 & 20.6 & 22.1 & 24.2 \\ 
B & 1B & 18.2 & 20.2 & 21.4 & 22.3 \\ 
H & 100B & 35.2 & 42.2 & 44.6 & 46.3 \\ 
H & 10B & 29.4 & 37.2 & 40.9 & 40.9 \\ 
H & 1B & 29 & 34.1 & 36.1 & 35.4 \\ 
L & 100B & 30.5 & 36.1 & 40.5 & 41.7 \\ 
L & 10B & 28.1 & 34.5 & 36.1 & 35.9 \\ 
L & 1B & 25.6 & 32.7 & 35.9 & 36.4 \\ 
\bottomrule
\end{tabular}
\end{table}

\begin{table}
\centering
\caption{Telugu Image-To-Text Zero-shot Retrieval.}
\begin{tabular}{ll|rrrr}
\toprule
Model & Data & 3B & 10B & 33B & 100B \\
\midrule
B & 100B & 5.7 & 10.3 & 15.6 & 19.5 \\ 
B & 10B & 6.6 & 9.4 & 13 & 12.9 \\ 
B & 1B & 2.4 & 2.1 & 2 & 1.9 \\ 
H & 100B & 7.5 & 17.5 & 27.4 & 34.7 \\ 
H & 10B & 7.9 & 16.2 & 23 & 26.9 \\ 
H & 1B & 3.8 & 4.2 & 3.4 & 3.6 \\ 
L & 100B & 7.2 & 16.1 & 23.8 & 30.3 \\ 
L & 10B & 7.2 & 14.9 & 19.4 & 23.3 \\ 
L & 1B & 4.5 & 4.7 & 4.2 & 3.9 \\ 
\bottomrule
\end{tabular}
\end{table}









\FloatBarrier

\section{Absolute Performance Levels}
\label{appendix:absolute_perf}

\begin{table}[!ht]
    \centering
    \caption{Crossmodal-3600 results spanning a wide performance spectrum.}
    \begin{tabular}{ll|rrrr}
    \toprule
        Range of Absolute Error Rate & Task & 10B & 100B & Error Rate Reduction \\ 
    \midrule
        0-20 & GeoDE: EastAsia & 5.01 & 4.68 & 0.33 \\ 
        0-20 & Caltech 10-shot Classification & 6.02 & 8.93 & -2.91 \\ 
        0-20 & GeoDE: Africa & 6.56 & 6.4 & 0.16 \\ 
        0-20 & Pet 0-shot Classification & 7.47 & 7.17 & 0.3 \\ 
        20-40 & Crossmodal-3600 Image-Text Retrieval: Turkish & 36.56 & 34.94 & 1.62 \\ 
        20-40 & Crossmodal-3600 Text-Image Retrieval: German & 36.56 & 36.99 & -0.43 \\ 
        20-40 & Crossmodal-3600 Image-Text Retrieval: Ukrainian & 36.94 & 33.25 & 3.69 \\ 
        20-40 & Crossmodal-3600 Image-Text Retrieval: Dutch & 38.06 & 37.44 & 0.62 \\ 
        40-60 & Crossmodal-3600 Image-Text Retrieval: Arabic & 41.6 & 41.0 & 0.6 \\ 
        40-60 & COCO Image-Text 0-shot Retrieval & 42.04 & 42.48 & -0.44 \\ 
        40-60 & Crossmodal-3600 Image-Text Retrieval: Japanese & 42.22 & 37.94 & 4.28 \\ 
        40-60 & Crossmodal-3600 Text-Image Retrieval: Russian & 43.58 & 43.08 & 0.5 \\ 
        60-80 & COCO Text-Image 0-shot Retrieval & 60.32 & 59.29 & 1.03 \\ 
        60-80 & Crossmodal-3600 Image-Text Retrieval: Bengali & 61.22 & 56.69 & 4.53 \\ 
        60-80 & Crossmodal-3600 Image-Text Retrieval: Hindi & 62.33 & 60.64 & 1.69 \\ 
        60-80 & Crossmodal-3600 Text-Image Retrieval: Greek & 65.45 & 64.1 & 1.35 \\ 
        80-100 & Crossmodal-3600 Text-Image Retrieval: Telugu & 84.44 & 79.57 & 4.87 \\ 
        80-100 & Crossmodal-3600 Text-Image Retrieval: Swahili & 87.47 & 85.67 & 1.8 \\ 
        80-100 & Crossmodal-3600 Image-Text Retrieval: Quechua & 93.06 & 92.78 & 0.28 \\ 
        80-100 & Crossmodal-3600 Image-Text Retrieval: Maori & 98.92 & 99.17 & -0.25 \\ 
    \bottomrule
    \end{tabular}
\end{table}

\end{document}